\documentclass{article}

\usepackage{arxiv}

\usepackage[utf8]{inputenc} 
\usepackage[T1]{fontenc}    
\usepackage{hyperref}       
\usepackage{url}            
\usepackage{booktabs}       
\usepackage{amsfonts}       
\usepackage{nicefrac}       
\usepackage{microtype}      
\usepackage{lipsum}
\usepackage{graphicx}
\graphicspath{ {./images/} }

\usepackage{graphicx}
\usepackage{hyperref}
\usepackage{lineno}

\usepackage{amssymb}
\usepackage{amsmath}

\usepackage[numbers]{natbib}

\usepackage{pgfplots}
\usepackage{xcolor}
\definecolor{lightbrown}{RGB}{181,101,29}
\usepgfplotslibrary{colormaps}

\pgfplotscreatecolormap{magma}{%
	rgb=(0.001462, 0.000466, 0.013866)
	rgb=(0.035520, 0.028397, 0.125209)
	rgb=(0.102815, 0.063010, 0.257854)
	rgb=(0.191460, 0.064818, 0.396152)
	rgb=(0.291366, 0.064553, 0.475462)
	rgb=(0.384299, 0.097855, 0.501002)
	rgb=(0.475780, 0.134577, 0.507921)
	rgb=(0.569172, 0.167454, 0.504105)
	rgb=(0.664915, 0.198075, 0.488836)
	rgb=(0.761077, 0.231214, 0.460162)
	rgb=(0.852126, 0.276106, 0.418573)
	rgb=(0.925937, 0.346844, 0.374959)
	rgb=(0.969680, 0.446936, 0.360311)
	rgb=(0.989363, 0.557873, 0.391671)
	rgb=(0.996580, 0.668256, 0.456192)
	rgb=(0.996727, 0.776795, 0.541039)
	rgb=(0.992440, 0.884330, 0.640099)
	rgb=(0.987053, 0.991438, 0.749504)
}
\pgfplotscreatecolormap{viridis}{%
	rgb=(0.267,0.00487,0.32942)
	rgb=(0.28192,0.08966,0.41241)
	rgb=(0.28026,0.1657,0.4765)
	rgb=(0.26366,0.23763,0.51877)
	rgb=(0.23744,0.3052,0.54192)
	rgb=(0.20862,0.36775,0.55267)
	rgb=(0.18225,0.42618,0.55711)
	rgb=(0.1592,0.48224,0.55807)
	rgb=(0.13777,0.53749,0.5549)
	rgb=(0.12115,0.59274,0.54465)
	rgb=(0.12808,0.64775,0.5235)
	rgb=(0.18065,0.7014,0.48819)
	rgb=(0.27415,0.75198,0.4366)
	rgb=(0.39517,0.79747,0.36775)
	rgb=(0.53561,0.83578,0.2819)
	rgb=(0.68895,0.86545,0.18272)
	rgb=(0.84557,0.88733,0.0997)
	rgb=(0.99324,0.90616,0.14394)
}

\pgfplotscreatecolormap{inferno}{%
	rgb=(0.001462, 0.000466, 0.013866)
	rgb=(0.037668, 0.025921, 0.132232)
	rgb=(0.116656, 0.047574, 0.272321)
	rgb=(0.217949, 0.036615, 0.383522)
	rgb=(0.316282, 0.053490, 0.425116)
	rgb=(0.410113, 0.087896, 0.433098)
	rgb=(0.503493, 0.121575, 0.423356)
	rgb=(0.596940, 0.154848, 0.398125)
	rgb=(0.688653, 0.192239, 0.357603)
	rgb=(0.775059, 0.239667, 0.303526)
	rgb=(0.851384, 0.302260, 0.239636)
	rgb=(0.912966, 0.381636, 0.169755)
	rgb=(0.956852, 0.475356, 0.094695)
	rgb=(0.981895, 0.579392, 0.026250)
	rgb=(0.987464, 0.690366, 0.079990)
	rgb=(0.973088, 0.805409, 0.216877)
	rgb=(0.947594, 0.917399, 0.410665)
	rgb=(0.988362, 0.998364, 0.644924)
}
\pgfplotscreatecolormap{plasma}{%
	rgb=(0.050383, 0.029803, 0.527975)
	rgb=(0.186213, 0.018803, 0.587228)
	rgb=(0.287076, 0.010855, 0.627295)
	rgb=(0.381047, 0.001814, 0.653068)
	rgb=(0.471457, 0.005678, 0.659897)
	rgb=(0.557243, 0.047331, 0.643443)
	rgb=(0.636008, 0.112092, 0.605205)
	rgb=(0.706178, 0.178437, 0.553657)
	rgb=(0.768090, 0.244817, 0.498465)
	rgb=(0.823132, 0.311261, 0.444806)
	rgb=(0.872303, 0.378774, 0.393355)
	rgb=(0.915471, 0.448807, 0.342890)
	rgb=(0.951344, 0.522850, 0.292275)
	rgb=(0.977856, 0.602051, 0.241387)
	rgb=(0.992541, 0.687030, 0.192170)
	rgb=(0.992505, 0.777967, 0.152855)
	rgb=(0.974443, 0.874622, 0.144061)
	rgb=(0.940015, 0.975158, 0.131326)
}

\pgfplotsset{
	colormap name=inferno,
}

\usepackage{tikz}
\usepackage{tikz-3dplot}
\usetikzlibrary{
	3d,
	perspective,
	calc,
	positioning,
	fit,
	backgrounds,
	math,
	arrows,
	external,
	tikzmark,
	shapes.geometric,
	decorations.markings,
	decorations.pathreplacing,
}

\tikzset{
	color of colormap = {200},
	/utils/exec={\xglobal\colorlet{cmapdark1}{.}},
	color of colormap = {300},
	/utils/exec={\xglobal\colorlet{cmapdark2}{.}},
	color of colormap = {400},
	/utils/exec={\xglobal\colorlet{cmapdark3}{.}},
	color of colormap = {500},
	/utils/exec={\xglobal\colorlet{cmapmed1}{.}},
	color of colormap = {575},
	/utils/exec={\xglobal\colorlet{cmapmed2}{.}},
	color of colormap = {650},
	/utils/exec={\xglobal\colorlet{cmapmed3}{.}},
	color of colormap = {725},
	/utils/exec={\xglobal\colorlet{cmaplight1}{.}},
	color of colormap = {800},
	/utils/exec={\xglobal\colorlet{cmaplight2}{.}},
	color of colormap = {900},
	/utils/exec={\xglobal\colorlet{cmaplight3}{.}},
	color of colormap = {0},
	black,
}

\usepackage{pgfplots}
\pgfplotsset{compat=1.17}
\usepgfplotslibrary{
	groupplots,
	colormaps,
	fillbetween,
}
\pgfdeclarelayer{background}
\pgfdeclarelayer{foreground}
\pgfsetlayers{background,main,foreground}

\usepackage{booktabs}
\usepackage{multirow}
\usepackage{adjustbox}
\usepackage{longtable}
\usepackage{csvsimple}

\usepackage{epsdice}
\usepackage{fontawesome}

\AtBeginDocument{%
  }

\title{Machine Learning Strategies for Parkinson Tremor Classification Using Wearable Sensor Data}

\author{
 Jesus Paucar-Escalante \\
  Institute of Computing\\
  State University of Campinas\\
  \texttt{j236865@dac.unicamp.br} \\
   \And
 Matheus Alves da Silva \\
  Faculty of Medical Sciences\\
  State University of Campinas\\
  \texttt{m200624@dac.unicamp.br} \\
  \And
 Bruno De Lima Sanches \\
  Institute of Physics\\
  State University of Campinas\\
  \texttt{delimabs@unicamp.br} \\
  \And
 Aurea Soriano-Vargas \\
 Departamento Académico de Ciencia de Computación y Datos \\
 Universidad de Ingenieria yTecnologia—UTEC, Lima, Peru \\
 \texttt{asoriano@utec.edu.pe}\\
  \And
 Laura Silveira Moriyama \\
 Faculty of Medical Sciences\\
 State University of Campinas\\
 \texttt{moriyama@unicamp.br}\\
 \And
 Esther Luna Colombini \\
 Institute of Computing\\
 State University of Campinas\\
 \texttt{estherlc@unicamp.br}
}

\begin{document}
\maketitle
\begin{abstract}
Parkinson\textquotesingle{}s disease (PD) is a neurological disorder requiring early and accurate diagnosis for effective management. Machine learning (ML) has emerged as a powerful tool to enhance PD classification and diagnostic accuracy, particularly by leveraging wearable sensor data. This survey comprehensively reviews current ML methodologies used in classifying Parkinsonian tremors, evaluating various tremor data acquisition methodologies, signal preprocessing techniques, and feature selection methods across time and frequency domains, highlighting practical approaches for tremor classification. The survey explores ML models utilized in existing studies, ranging from traditional methods such as Support Vector Machines (SVM) and Random Forests to advanced deep learning architectures like Convolutional Neural Networks (CNN) and Long Short-Term Memory networks (LSTM). We assess the efficacy of these models in classifying tremor patterns associated with PD, considering their strengths and limitations. Furthermore, we discuss challenges and discrepancies in current research and broader challenges in applying ML to PD diagnosis using wearable sensor data. We also outline future research directions to advance ML applications in PD diagnostics, providing insights for researchers and practitioners.
\end{abstract}


\section{Introduction}
\label{sec:introduction}

Parkinson's disease (PD) is the second most common neurodegenerative condition in the world. Some of the motor symptoms are caused by low levels of dopamine secondary to neurodegenerative changes and neuronal death in the substantia nigra. The finding of bradykinesia (slowness of movement and decrement in amplitude or speed as movements are continued) \cite{postuma2015mds} is necessary for the diagnosis, but a large number of patients also present tremor as a characteristic feature \cite{ruiz2011initial}, causing severe impairments in the patient's life quality \cite{muslimovic2008determinants}. Idiopathic Parkinson's disease primarily affects the elderly, but it can be diagnosed in adults under 40 \cite{quinn1987young}. The situation becomes even more severe when considering that, despite advances in symptomatic treatments, there is still no intervention that cures or changes the natural course of the disease \cite{capriotti2016parkinson}. 

However, early diagnosis is one of the major challenges of PD: it is estimated that more than 60\% of the dopaminergic neurons that play a key role in PD's development have already been lost when the first clinical symptoms that catch the attention are observed before the motor symptoms are clinically recognized \cite{cheng2010clinical}. The diagnosis is based on the patient's history and symptomatology, combined with physical and neurological examinations \cite{armstrong2020diagnosis}. Over the past decades, biomarkers have been proposed for PD diagnosis, which can usually be classified into four types: clinical, imaging, biochemical, and genetic \cite{emamzadeh2018parkinson}. Despite progress, the search for biomarkers capable of tracking the onset of symptoms in an early stage of the disease is still an unresolved issue, and advancements in this area - even if marginal - can impact initiating symptomatic therapy quickly. The search for biomarkers for tracking disease progression is also essential, which may help the development of disease-modifying treatments \cite{mollenhauer2023status}, thereby delaying the disease's impairments. 

Despite the motor symptoms being the primary hallmark of this neurological condition, the impact on patient's daily lives extends beyond motor involvement. Neuropsychiatric symptoms include mood, anxiety, psychosis, impulsivity, and compulsivity \cite{burchill2024impact}. Gastrointestinal symptoms include constipation (probably the earliest symptoms in many patients), slowed gastric emptying, bloating, and cramps \cite{pasricha2024management}. Sleep disturbances include insomnia, daytime sleepiness, and REM-sleep behavior disorder \cite{dodet2024sleep}. Dysautonomia with orthostatic hypotension, pain, skin conditions, olfactory deficit, and many other symptoms are characteristic of the disease. Non-motor symptoms are currently the primary determinant of life quality in PD \cite{bock2022association}.

Most therapies target the movement-related motor symptoms (bradykinesia, tremor, rigidity, and gait/posture abnormalities) but cause or worsen the non-motor symptoms. These include constipation, sleep disturbances, neuropsychiatric features (like mood disorders, psychosis, impulse control disorders), and dysautonomia \cite{palma2018treatment}. Therefore, appropriate monitoring of motor symptoms to avoid over-treatment is paramount to the life quality of PD patients. Currently, the tracking of PD symptoms is predominantly clinical and is conducted through regular consultations with a physician. In the research setting, objective measurements are largely performed using scales, such as the Unified Parkinson's Disease Rating Scale (UPDRS) and the Hoehn and Yahr Scale (HY), which are widely utilized, especially the versions revised by the Movement Disorder Society \cite{goetz2004movement, goetz2008movement}. The UPDRS aims to assess various aspects of the patient's condition, including motor and non-motor symptoms in their daily life and motor complications of the disease; the assessment consists of different parts involving clinical and self-report questionnaires. The HY scale classifies the disease's progression into stages ranging from 1 to 5 and is included within the UPDRS. Many specific scales are used for various symptoms (neuropsychiatric, sleep, pain, and so forth). Both motor and non-motor symptoms fluctuate over time; therefore, patient diaries are also used to record these fluctuations. These diaries often lack precision, especially in patients with cognitive impairment \cite{goetz1997efficacy}.

Measuring motor symptoms quantitatively and objectively can be a powerful and viable alternative, especially with the advancement of technology. The use of wearable devices for continuous monitoring during daily activities could support approaches to quantify motor symptoms during their daily activities outside the clinic, similar to how a Holter monitor evaluates cardiac function over a relatively long period. Accelerometers and gyroscopes in commercial smartwatches can provide precise measurements of movement direction, frequency, and intensity, particularly stride length, which is reduced in most PD patients \cite{sayeed2015adapted}. AI-driven medical technologies are swiftly progressing into viable options for detecting and managing diseases \cite{briganti2020artificial} as the volume of data from diverse sensors grows. AI algorithms can adjust and offer enhanced autonomy, resulting in more individualized treatment options. Regular and frequent data collection allows for creating daily symptom profiles that can assist in developing an ideal therapeutic regimen \cite{powers2021smartwatch}. 

Studies of objective monitoring of PD symptoms typically feature a constrained sample size, often fewer than 100 participants, and exhibit limited diversity in age, gender, and ethnicity. These constraints hinder the broader applicability of the findings. Therefore, while there are still some limitations to the routine application of monitoring with intelligent wearable sensors, the application is gaining ground from a research perspective, given its evident potential.

\subsection{Contributions}
\label{sec:contributions}
This survey makes the following key contributions:

\begin{itemize}
    \item \textbf{Data Acquisition Methodologies}: Evaluate and compare the methodologies used to acquire tremor data, including body location and sampling frequency. 
    \item \textbf{Signal Preprocessing}: Analyze preprocessing techniques applied to tremor signals before input to the model, such as filtering, normalization, and artifact removal. 
    \item \textbf{Feature Selection}: Review and compare approaches used for feature selection in both time and frequency domains. 
    \item \textbf{Machine Learning and Deep Learning Models}: Examine specific models used in each study, highlighting both traditional methods (SVM, Random Forest, etc.) and deep learning models (CNN, LSTM, etc.) 
    \item \textbf{Frequency Bands and Comparison with Physiological References}: Investigate specific frequency bands considered in each study for tremor classification. 
    \item \textbf{Model Explainability}: Evaluate the presence of explainability techniques in proposed models to understand how classification decisions are made. 
    \item \textbf{Contradictions and Challenges}: Highlight inconsistencies between studies, especially regarding contradictory results or divergent methodological approaches. Identify common challenges faced by reviewed studies and propose possible solutions. 
\end{itemize}

The subsequent sections of this study are organized as follows. In Section \ref{sec:background}, we present concepts about Parkinson's disease, focusing specifically on ``tremor''. Section \ref{sec:taxonomy} introduces the proposed taxonomy and describes the structure of each classification. In Section \ref{sec:section4}, we examine each classification within our proposed taxonomy and analyze their merits and shortcomings, providing an overview of unresolved issues within the field and suggesting potential avenues for future research. Lastly, Section \ref{sec:conclusion} encapsulates concluding remarks.

\section{Background}
\label{sec:background}

This section presents an overview of Parkinson's disease (PD) characterization, providing the background to understand the importance and nature of tremors in PD. Subsequently, we explore potential avenues for contributions from the computational domain in the context of Parkinson's disease research.

\subsection{Parkinson Disease}

Parkinson Disease (PD) is a chronic and progressive neurodegenerative disorder characterized by a spectrum of motor symptoms, including bradykinesia, rest tremor, rigidity, and postural instability, and non-motor symptoms, combined with late-onset motor manifestations such as postural instability, falls, freezing of gait, and speech and swallowing difficulties, lead to challenges to healthcare providers managing the disease over extended periods \cite{world2006neurological, ruiz2011initial}.

The precise PD etiology and progression remain elusive. Ongoing research focuses on enhancing our comprehension of this condition. Nevertheless, PD exhibits a substantial global prevalence. According to World Health Organization (WHO) data, PD ranks as the second most prevalent neurological disorder worldwide. It has displayed an increase in disability and mortality rates, with reported cases doubling over the past quarter-century, reaching $8.5$ million globally by 2019 (approximately $2\%$ of the global population) \cite{whoParkinson2021, world2006neurological, arredondo2018breve}. Current estimates indicate that PD was responsible for $5.8$ million disability-adjusted life years and contributed to $329,000$ fatalities in the same year. These statistics reflect a notable $100\%$ upsurge in mortality rates compared to 2000 \cite{arredondo2018breve}.

PD is a multifaceted challenge that impacts the patient and their environment in various ways. Research endeavors focus on enhancing the quality of life of individuals affected by PD, emphasizing the need to address symptoms once the origins and progression of the disease become more comprehensible. As described in the introduction, PD causes both motor and non-motor symptoms, and bradykinesia, tremor, rigidity, and gait abnormalities are among the first to be used to define and diagnose the condition.

\subsection{Pathophysiology of Parkinson's disease}

The primary pathological hallmarks of Parkinson's disease (PD) involve the loss of dopaminergic neurons, resulting in the depigmentation of the substantia nigra pars compacta (SNpc), and the frequent presence of intracytoplasmic inclusions known as Lewy bodies (LBs), which are identifiable by a peripheral pale halo in a neuropathological examination of the nervous system \cite{balestrino2020parkinson,ball2019parkinson, beitz2014parkinson}. Although the dopaminergic loss is still considered a main culprit of bradykinesia, it is also known that various anatomical and network changes are involved in PD pathophysiology of motor manifestations of disease \cite{cabreira2019doencca}, and include that tremor generation, in particular, consists of the cerebellum, thalamus, and cortical network \cite{duval2016brain}. One of the models proposed is that tremors, mainly rest tremors, stem from abnormal activity within the basal ganglia (BG), initiated by the thalamus and further potentiated by the cerebellum, interpreting the tremor as a voluntary movement. Neurotransmitter changes are not limited to dopamine but involve various neurotransmitter systems in tremor generation \cite{dirkx2022pathophysiology}. Genetic investigations have elucidated a link between PD and more than $11$ distinct genes, each with varying functions, showing a complex background for idiopathic Parkinson's disease. While some monogenetic inheritance patterns were discovered, most of the patients living with Parkinson's disease have a complex multifactorial disorder in which stochastic, environmental, and genetic factors interact. 

\subsection{Symptoms}

Idiopathic Parkinson's Disease is a heterogeneous condition. It affects each patient living with the disorder with different motor and non-motor symptoms. Besides the PD diagnosis, when there is clinical evidence of Parkinsonism, many non-motor symptoms may precede the appearance of motor symptoms.

\subsubsection{Motor symptoms}

The primary motor features of PD include bradykinesia, tremors (mainly rest tremors), and rigidity. Other symptoms may develop as the disease progresses \cite{moustafa2016motor}. The Movement Disorder Society defines bradykinesia as the slowness of movement and decrement in amplitude or speed (or progressive hesitations/halts) as movements are continued. Rigidity is defined as the low passive movement of major joints with the patient in a relaxed position and the examiner manipulating the limbs and neck, presenting with a velocity-independent feature (one of the clinical clues for differentiation with spasticity). 

Parkinson's disease patients may present with action, postural, or rest tremor, but for the diagnosis of Parkinson's disease, the rest tremor is more common. From the clinical standpoint, tremor is expected to have a 4 - 6 Hz frequency and is suppressed completely or partially during movement. Some patients may have characteristic re-emergent tremor \cite{postuma2015mds}. Other common symptoms are postural instability and axial abnormalities such as camptocormia and Pisa syndrome, gait disturbances such as freezing, eye movement disturbances, and postural instability \cite{balestrino2020parkinson, cabreira2019doencca}

\subsubsection{Non-Motor symptoms} 

There are many nonmotor symptoms associated with PD. Some may even develop before the motor symptoms, like hyposmia and constipation. Nevertheless, these motor symptoms are highly heterogeneous among the people living with Parkinson's Disease and can be so or more debilitating than the motor symptoms \cite{elsworth2020parkinson, khatri2020anxiety}. In general, as PD progresses, the development of both motor and non-motor symptoms can result in partial or complete disability, contingent on the evolution and severity of the disease. Even in the advanced stages, hospitalization and pharmacological interventions are often necessary to address physical symptoms such as tremors and bradykinesia. It is crucial not to overlook the psychological aspects, as the progression of motor symptoms and their challenges can lead to psychological issues such as depression and anxiety, further complicating the overall well-being and quality of life of individuals living with PD \cite{khatri2020anxiety, marsh2013depression}. 

These multifaceted challenges underscore the importance of comprehensive and holistic care for PD patients, addressing their physical and emotional needs. The context of non-motor symptoms associated with PD is vast and encompasses many manifestations. However, a challenge lies in directly attributing these symptoms to PD, as many are shared with other medical conditions. As a result, it is uncommon to make a direct and definitive diagnosis of PD based solely on non-motor symptoms. 

Currently, the standard methods for detecting and diagnosing PD primarily rely on neurological evaluation of motor symptoms because the presence of Parkinsonism is necessary for the diagnosis. Imaging techniques such as Magnetic Resonance Imaging (MRI) and Single-photon emission computed tomography (SPECT) with a dopaminergic tracer may be used as ancillary test \cite{bergamino2020assessing}. Although PD is currently diagnosed through clinical evaluation, noninvasive characterization monitoring must be used to differentiate diagnoses better and monitor clinical response, mainly focusing on recognizing motor and non-motor symptoms.

\subsection{Detection of tremors}

\emph{Tremors} in PD are characterized by involuntary, rhythmic, oscillatory movements of a specific body part, often affecting the upper extremities \cite{saifee2019tremor}. Tremor may be classified as in 18 clinical categories \cite{bhatia2018consensus}, of which three are more common:

\begin{enumerate}
    \item \textbf{Rest Tremor:} This type of tremor occurs when a limb is inactive and only influenced by gravity. Involuntary oscillatory movements characterize it while the limb is at rest.
    \item \textbf{Action Tremor:}  Action tremor occurs when involuntary, pronounced oscillatory movements accompany a voluntary movement. Muscle contractions play a role in action tremors. Action tremors may be divided into kinetic, postural, and isometric tremors. Postural tremor manifests during activities against gravity. PD tremor may have a characteristic re-emergent tremor.
    \item \textbf{Postural Tremor:} Postural tremor manifests during activities that require maintaining a position against gravity, such as standing upright. It involves involuntary oscillatory movements that occur when opposing gravity's force.
\end{enumerate}

To characterize tremors effectively, measuring devices must be capable of capturing various movement characteristics, including direction, frequency, intensity, and stride length. Portable inertial devices, such as accelerometers and gyroscopes, have become valuable tools for monitoring PD motor symptoms. These sensors have been incorporated into wearable devices like watches and smartphones, offering sufficient accuracy to capture the nuances of tremor movements. This technology presents a promising option for continuous and objective monitoring of tremors in PD patients, aiding in assessing symptom severity and treatment effectiveness \cite{sayeed2015adapted}.

Symptoms are presented in a particular way in each patient, forcing us to characterize the tremors based on a global characteristic to differentiate and measure the degree according to the presence of the same. In this sense, one of the features used at the pathophysiological level is the frequency (measured in Hz, see Table \ref{tab:Table1}.

\begin{table}[ht]
\scriptsize
\resizebox{\columnwidth}{!}{%
\begin{tabular}{|c|c|c|c|l}
\cline{1-4}
\textbf{Tremor}                               & \textbf{Body part commonly affected}          & \textbf{Type of tremor}       & \textbf{Frequency} &  \\ \cline{1-4}
\emph{Essential tremor}                       & Bilateral upper limb                          & Action                        & 4 - 12 Hz          &  \\ \cline{1-4}
\emph{Enhanced physiological tremor}          & Bilateral upper limb                          & Postural \textgreater Kinetic & 8 - 12 Hz          &  \\ \cline{1-4}
\emph{Dystonic tremor}                        & Asymmetric upper limbs and head               & Action, occasionally rest     & 3 - 5 Hz           &  \\ \cline{1-4}
\emph{Cereberall tremor}                      & Asymmetric or symmetric limbs, head and trunk & Intention                     & \textless 5 Hz     &  \\ \cline{1-4}
\emph{Holmes tremor}                          & Unilateral upper limb                         & Kinetic \textgreater Rest     & \textless 5 Hz     &  \\ \cline{1-4}
\emph{Palatal tremor}                         & Palate (oculofacial structures)               & Rest                          & \textless 3 Hz     &  \\ \cline{1-4}
\emph{Parkinsonian tremor (common)}           & Asymmetric upper limbs, legs, jaw             & Rest, re-emergent             & 3 - 6 Hz           &  \\ \cline{1-4}
\emph{Neuropathic tremor}                     & Bilateral upper limbs                         & Action                        & 4 - 8 Hz           &  \\ \cline{1-4}
\emph{Orthostatic tremor}                     & Bilateral lower limbs                         & Isometric on standing         & 13 - 18 Hz         &  \\ \cline{1-4}
\emph{Functional tremor}                      & Upper limbs                                   & All states                    & 3 - 6 Hz           &  \\ \cline{1-4}
\emph{Drug induced}                           & Bilateral upper limbs                         & Action \textgreater Rest      & 8 - 12 Hz          &  \\ \cline{1-4}
\end{tabular}%
}
\caption{Tremor characteristics table. Adapted from Saifee (2019) \cite{saifee2019tremor}.}
\label{tab:Table1}
\end{table}

While analyzing the temporal aspects of tremor signals remains important, the frequency characteristics add a specific dimension that can aid in distinguishing between different types of tremors. Frequency analysis can provide valuable insights into tremors' nature and underlying mechanisms. Some conditions of the patient may, however, complicate the evaluation of frequency, and the review must be associated with the amplitude of the tremor, the direction of the tremor (flexion-extension, pronation-supination or mixed), the body distribution of the tremor, and which position the tremor is more evidenced.

In recent years, the research efforts toward signal processing techniques have evolved. They often utilize advancements in artificial intelligence with and without specific frequency evaluation. These approaches can potentially enhance the accuracy and efficiency of tremor analysis and monitoring \cite{saifee2019tremor}. A recent literature review of 24 studies \cite{ancona2022wearables} highlights the potential of artificial intelligence (AI) solutions integrated with inertial devices for continuous tremor monitoring. While this review underscores the opportunities in using AI for signal capture in home environments, it lacks a detailed exploration of the methodologies and intelligent models employed in each study. Moreover, many of the reviewed studies involve small patient cohorts, raising concerns about the generalizability of their findings. Further research and larger-scale studies are needed to establish more comprehensive and robust conclusions within this field.

In contrast, our study distinguishes itself by thoroughly examining the methodologies and AI models underlying each investigation. This approach addresses the gaps identified in previous literature and aims to develop more robust and scalable AI-driven solutions in tremor monitoring.

\section{Taxonomy}
\label{sec:taxonomy}

The analysis of tremors plays a pivotal role in monitoring PD. In this context, the current section introduces a taxonomy structured into several critical stages, facilitating performance measurement for the effective understanding of detecting motor patterns associated with PD.

\begin{figure}
    \centering
    \input{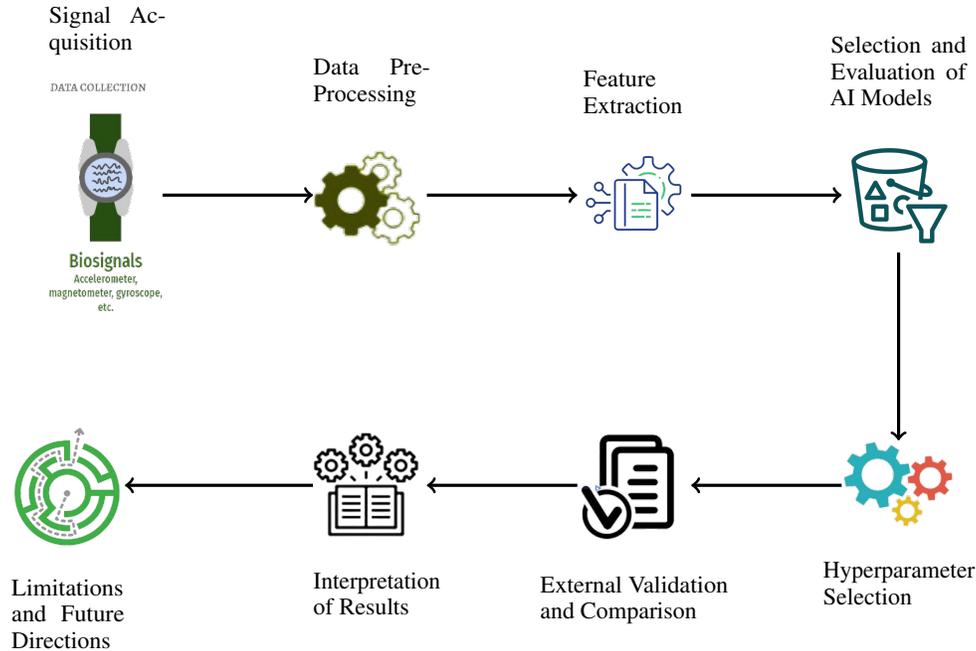}
    \caption{%
        Illustration of the proposed general structure (Taxonomy).
    }%
    \label{fig:figure1}

\end{figure}

Considerations within the proposed framework encompass features for signal capture, signal processing, and results analysis for the artificial intelligence models employed. The outlined structure is illustrated in Figure \ref{fig:figure1}, delineating eight stages wherein specific points will be scrutinized across all selected studies. These stages and points are expounded upon in the subsequent discussion.

To systematically discuss the algorithms within this framework, we reference the study by Skandha et al. (2020) \cite{skandha20203}, categorizing algorithms into three generations, thereby highlighting their evolution over time. The first generation comprises manual or Gaussian Shell Map (GSM-based techniques), characterized by practical, rule-based problem-solving methods. The second generation introduces machine learning (ML)-based techniques, where models utilize algorithms to analyze data and generate insights, marking a shift towards a more automated and data-driven approach. The third generation, representing the most advanced stage, involves deep learning (DL) methods. This generation leverages complex neural networks to learn intricate patterns and representations from data, enabling high-level abstraction and sophisticated feature extraction.

\textbf{Signal Acquisition}.  First, we analyze the existing distribution within the dataset to observe the variability of the classes present. The type of sensor used and the sampling frequency with which the signal is captured in each study are identified. These general characteristics provide information about the dataset's composition and the signals' constitution, directly influencing part of the second stage corresponding to data preprocessing.

\textbf{Data Preprocessing}. In the second stage, the filtering bands used for each study are analyzed, observing existing correlations with the addressed problem and checking if they align with theoretical physiological bands. Similarly, information about alternative processing, such as normalization, segmentation, and/or signal transformation, is extracted through an artificial intelligence network before a classification stage.

\textbf{Feature Extraction}. In the third stage, feature extraction from a context of complete, segmented, or image signals is analyzed within a sub-classification of temporality, statistics, or frequency. The aim is to correlate similar features among studies that may reveal significance in the characterization of tremors.

\textbf{Selection and Evaluations of AI models}. In the fourth stage, the architectures employed are examined through the lens of the sub-classification proposed by Skandha et al. (2020) \cite{skandha20203}, as previously discussed. This approach allows for assessing the study's temporality and the specific metrics used to evaluate the reported results within each study.

\textbf{Hyper-parameter Selection}. In the fifth stage, the parameters used for each model are addressed, along with the selection method, and/or these parameters are searched for whenever they are reported within the study.

\textbf{External Validation and Comparison}. In the sixth stage, model validation is addressed by identifying the comparison of reported results with previous studies addressing the same problem and whose comparison can be conclusive from a technical perspective.

\textbf{Interpretation of results}. In the seventh stage, the results obtained in each study are analyzed, emphasizing the approach considered in each study, particularly if such an approach involves implications in the context and/or clinical practice.

\textbf{Limitations and Future Directions}. In the eighth and final stage, the identification and discussion of the limitations of the studies are addressed, along with future suggestions that can be considered in future research.

\section{Studies Review}
\label{sec:section4}

On July 3, 2023, a comprehensive search was conducted across five major databases: ACM, PubMed, Scopus, Web of Science, and IEEE Xplore. The search spanned a five-year period, from 2018 to 2023, including all studies published up to the search date that met the established inclusion and exclusion criteria. The inclusion criteria focused on studies that characterized Parkinson's disease, employed artificial intelligence models, recognized motor symptoms associated with Parkinson's, and utilized inertial sensors. Studies were excluded if they were duplicate citations, lacked full-text availability as of the search date, or were focused on biochemical analysis or genetic characteristics. Following these criteria, 34 studies were identified and will be discussed in the subsequent sections.

In the following section, we will systematically examine the characteristics of the selected studies within the proposed taxonomy. This analysis will provide a detailed view of the overall composition of the dataset and the signals it encompasses. Subsequently, we will describe the classification model and interpret the results. To establish a temporal perspective of the studies' development, we initially present general statistics regarding the chronology, focusing on the detection and/or identification of tremors as a central theme.

\begin{figure}[ht] 
\centering 
\includegraphics[width=0.65\columnwidth]{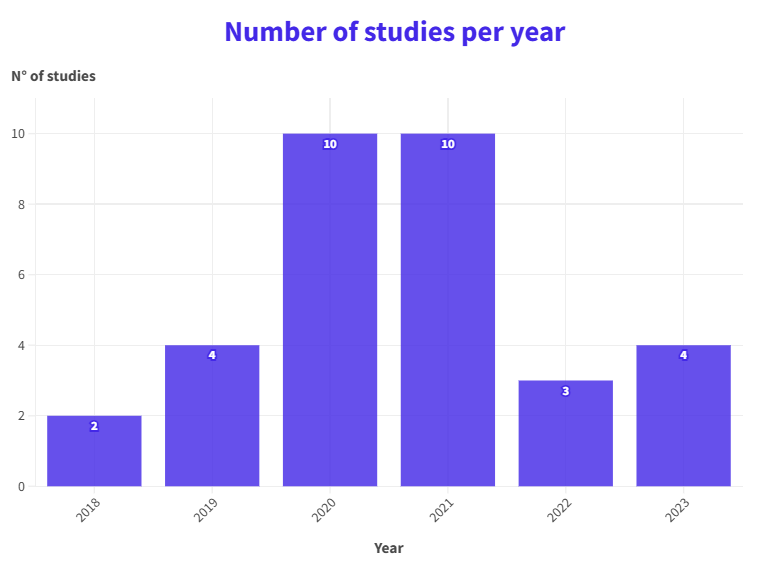}
\caption{Number of studies per year.}
\label{fig:figure2} 
\end{figure}

As a preliminary summary, Figure \ref{fig:figure2} illustrates the number of studies per year from 2018 to 2023. It highlights the predominance of studies in 2020 and 2021, with ten studies each, outnumbering the other years. Also, a general upward trend is observed from 2018 onwards, with a slight decrease in 2021, giving an idea of the evolution of interest in addressing the above problem during the specified period.

\begin{figure}
    \centering
    \resizebox{!}{0.7\textheight}{%
        \input{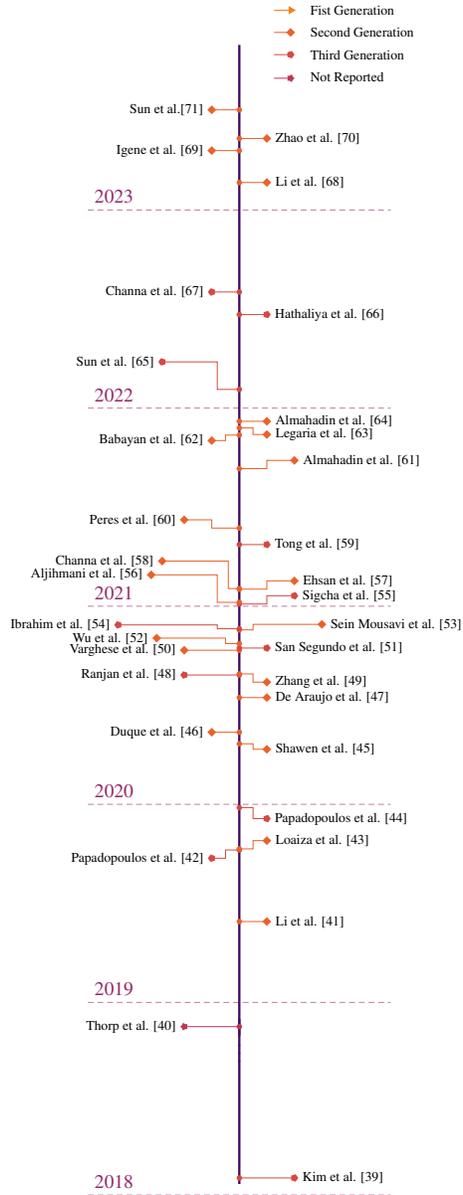}
        }
    \caption{%
        Chronology illustrating studies aimed at Parkinson's classification using inertial sensors from 2018 to the end of 2023.
		The timeline includes a sub-classification dependent on the generation to which each architecture used in each study belongs, taking as a reference the classification of Skandlha et al. in \cite{skandha20203}.
		The dates shown in the timeline are the reference dates of publication in their respective journals. The corresponding citation is in numerical form.
    }%
    \label{fig:figure3}
\end{figure}

The studies are presented in chronological order as a secondary summary. Our research, covering a 5-year period from 2018 to 2023, is illustrated in Figure \ref{fig:figure3}. This figure displays a general timeline, segmented by reference lines indicating each development year, showing each study's chronological order and algorithm generation. The aim is to relate the temporal progression of the studies and the trend in algorithm usage over time.

\begin{figure}[ht] 
\centering 
\includegraphics[width=0.7\columnwidth]{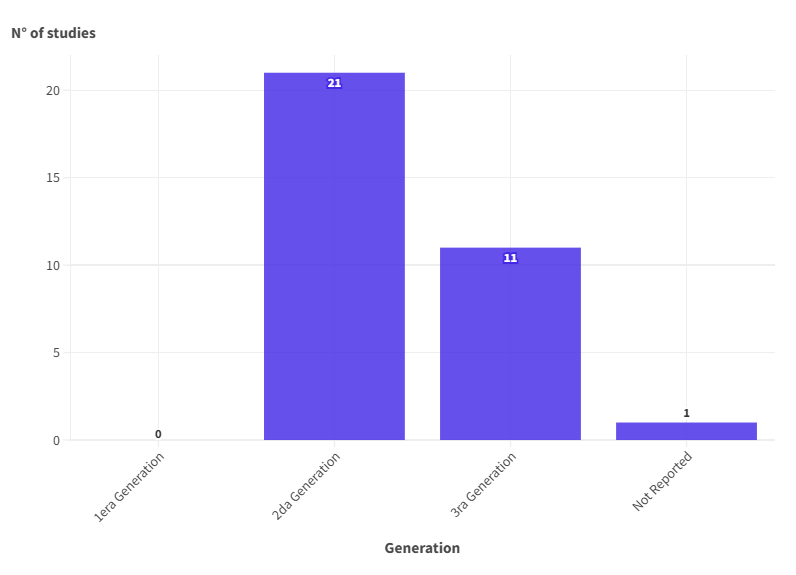}
\caption{Distribution per generation.}
\label{fig:figure4}
\end{figure}

This initial generalized approximation of algorithm generations is represented in Figure \ref{fig:figure4}, revealing the distribution and prevalence of specific algorithms for tremor classification, which is not necessarily associated with the effectiveness of each one of them, so it should be taken as a reference for counting studies.

Some general characteristics associated with the collected studies have been discussed. However, before moving specifically to the analysis of each characteristic, two important points should be made; the first point will be the existing public datasets used in some specific articles, and the second point will show a summary of all the important characteristics of the collected studies.

Regarding the datasets used, five public datasets used in 9 studies have been identified, and private datasets have been used for the other 24 studies. In Table \ref{tab:Dataset}, we briefly describe the public datasets to keep the reference of the main characteristics that encompass them.

\begin{table}[ht]
\centering
\scriptsize
\caption{Public dataset}
\label{tab:Dataset}
\begin{tabular}{|l|l|l|}
\hline
\textbf{Name of Dataset}        & \textbf{Link of dataset} & \textbf{Articles used} \\ \hline
\textit{PD-BioStampRC21}\cite{biostamp}        & 
\href{https://ieee-dataport.org/open-access/pd-biostamprc21-parkinsons-disease-accelerometry-dataset-five-wearable-sensor-study-0}{BioStamp}  &                       \cite{igene2023machine, hathaliya2022deep}
\\ 
\hline

\textit{MJFF Levodopa}\cite{daneault2021accelerometer}          & 
\href{https://www.synapse.org/#!Synapse:syn20681023/wiki/594678}{Levodopa}         &                        
\cite{li2023improved, channa2022parkinson, almahadin2021parkinson, almahadin2021task}
\\ 
\hline

\textit{Parkinson Desease Data}\cite{zhang2020personalized}     & 
\href{http://www.humansensing.cs.cmu.edu/pdd/}{PPD}                             &                     
\cite{zhang2020comparing}
\\ 
\hline

\textit{IMU Parkinson Dataset}\cite{alexandrosp}            & 
\href{https://zenodo.org/record/3519213}{IMU Zenodo}                      &                        
\cite{papadopoulos2019detecting}
\\ 
\hline

\textit{HandPD Dataset}\cite{PereiraSIBGRAPI:16}         & 
\href{https://wwwp.fc.unesp.br/~papa/pub/datasets/Handpd/}{HandPD}                          &                        
\cite{papadopoulos2019multiple}
\\ 
\hline
\end{tabular}
\end{table}

On the other hand, Table \ref{tab:table_all} summarizes the information from the 33 studies to comment on essential characteristics that will be discussed in the following section.

\begin{table*}[!htp]
	\begin{center}
		\caption{
			This table provides a detailed comparison of the studies collected. It classifies studies according to key attributes, such as year of publication, type of dataset (public or private), sensors used (accelerometer, gyroscope, magnetometer), generation of the model applied (first, second, or third generation), presence of a control group, proportion of healthy and unhealthy participants in the dataset, and sampling frequency. Icons are used to visually represent different types of sensors and data sets, while the distribution of participants' health status is represented by bar charts showing the percentage distribution. This overview allows easy comparison of methodologies and technologies across studies, highlighting trends and differences in sensor use, data accessibility, and model development.
		}%
		\label{tab:table_all}
		{
\newcommand{\tabheader}{Study &Year &Dataset &Type Sensor &Generation &Group Control &Ratio Healthy/No Healthy &Sampling Frequency \\}


\newcommand{\publicdata}{\color{cmapdark1}\faUnlock\color{black}}
\newcommand{\privatedata}{\color{cmapdark1}\faLock\color{black}}

\newcommand{\acc}{
    \tikzexternaldisable
    \begin{tikzpicture}[scale=0.045]
        \draw[thick, cmapdark2] (0,0) circle(3);
        
        \filldraw[fill=cmapdark2, thick] (-0.5,-0.5) rectangle (0.5,0.5);

        \draw[thick, cmapdark2, ->] (0, 0.5) -- (0, 2.5); 
        \draw[thick, cmapdark2, ->] (0.5, 0) -- (2.5, 0); 
        \draw[thick, cmapdark2, ->] (0, -0.5) -- (0, -2.5); 
        \draw[thick, cmapdark2, ->] (-0.5, 0) -- (-2.5, 0); 
    
        \draw[thick, cmapdark2] (2, 1.5) arc[start angle=0, end angle=40, radius=2.5];
        \draw[thick, cmapdark2] (2, 1.2) arc[start angle=0, end angle=40, radius=2.0];
        \draw[thick, cmapdark2] (2, 0.9) arc[start angle=0, end angle=40, radius=1.5];
    \end{tikzpicture}
    \tikzexternalenable
}

\newcommand{\gyr}{
    \tikzexternaldisable
    \begin{tikzpicture}[scale=0.15]
        \draw[thick, color=cmapdark2] (0,0) circle(1);
        \draw[thick, color=cmapdark2] (0,0) ellipse (0.7 and 0.3);
        \draw[thick, color=cmapdark2, rotate=45] (0,0) ellipse (0.7 and 0.3);
        \draw[thick, color=cmapdark2, rotate=-45] (0,0) ellipse (0.7 and 0.3);
        \filldraw[color=cmapdark2] (0,0) circle(0.1);
        \filldraw[color=cmapdark2] (-1,0) circle(0.1);
        \filldraw[color=cmapdark2] (1,0) circle(0.1);
    \end{tikzpicture}
    \tikzexternalenable
}

\newcommand{\mymagnetometericon}{
    \tikzexternaldisable
    \begin{tikzpicture}[scale=0.045]
        \filldraw[thick, cmapdark2] (0,0) circle(3);

        \filldraw[fill=white, thick] (0,0) circle(2.5);

        \filldraw[fill=cmapdark2, thick] (0,0) circle(0.2);

        \draw[thick, cmapdark2, fill=gray] (0,0) -- (1.8,1.8) -- (0,0.5) -- (-1.8,-1.8) -- cycle;
    
        \foreach \angle in {45, 135, 225, 315} {
            \draw[thick, cmapdark2] (\angle:2.5) -- (\angle:3);
        }

        \draw[thick, cmapdark2] (0,3) arc[start angle=90, end angle=135, radius=3];
        \draw[thick, cmapdark2] (-3,0) arc[start angle=180, end angle=225, radius=3];
    \end{tikzpicture}
    \tikzexternalenable
}

\newcommand{\notr}{\color{cmapdark2}\faQuestionCircle\color{black}}

\newcommand{\firstgen}{
    \tikzexternaldisable
    \begin{tikzpicture}[scale=0.15]
        \filldraw[color=cmapdark2, fill=white] (0,0) circle(1);
        \node[color=cmapdark2] at (0,0) {\textbf{1}};
    \end{tikzpicture}
    \tikzexternalenable
}

\newcommand{\secondgen}{
    \tikzexternaldisable
    \begin{tikzpicture}[scale=0.15]
        \filldraw[color=cmapdark2, fill=white] (0,0) circle(1);
        \node[color=cmapdark2] at (0,0) {\textbf{2}};
    \end{tikzpicture}
    \tikzexternalenable
}

\newcommand{\thirdgen}{
    \tikzexternaldisable
    \begin{tikzpicture}[scale=0.15]
        \filldraw[color=cmapdark2, fill=white] (0,0) circle(1);
        \node[color=cmapdark2] at (0,0) {\textbf{3}};
    \end{tikzpicture}
    \tikzexternalenable
}

\newcommand{\chk}{\color{cmapdark3} \checkmark}

\newcommand{\xmark}{
    \tikzexternaldisable
    \begin{tikzpicture}[scale=0.3]
        \draw[thick, color=cmapdark3] (0,0) -- (0.6,0.6);
        \draw[thick, color=cmapdark3] (0,0.6) -- (0.6,0);
    \end{tikzpicture}
    \tikzexternalenable
}

\newcommand{\distributionbar}[2]{
    \tikzexternaldisable
    \begin{tikzpicture}[scale=0.3, baseline]
    
        \filldraw[fill=gray!30] (0,0) rectangle (1,0.5);
        
        \filldraw[fill=cmapdark2] (0,0) rectangle (#1/100,0.5);
        
        \draw (0,0) rectangle (1,0.5);
        
        \node[anchor=west, font=\scriptsize] at (1.1,0.25) {#1\% - #2\%};
    \end{tikzpicture}
    \tikzexternalenable
}

\begin{adjustbox}{width=0.98\linewidth}
    \begin{tabular}{l ccccccc}
        \toprule
        \tabheader
        \midrule
        \multirow{1}{*}{Kim et al. \cite{kim2018wrist}} &2018 &\privatedata &\acc/\gyr &\thirdgen &\xmark &\distributionbar{0}{100} &\text{125Hz} \\
        \multirow{1}{*}{Thorp et al. \cite{thorp2018monitoring}} &2018 &\privatedata &\acc &\notr &\notr &\notr &\notr \\
        \midrule
        \multirow{1}{*}{Li et al. \cite{li2019monitoring}} &2019 &\privatedata &\acc/\gyr &\secondgen &\chk &\distributionbar{48}{52} &\text{40Hz} \\
        \multirow{1}{*}{Papadopoulos et al. \cite{papadopoulos2019multiple}} &2019 &\privatedata &\notr &\thirdgen &\notr &\notr &\notr \\
        \multirow{1}{*}{Loaiza et al. \cite{loaiza2019using}} &2019 &\privatedata &\acc &\secondgen &\chk &\distributionbar{23.08}{76.92} &\text{100Hz} \\
        \multirow{1}{*}{Papadopoulos et al. \cite{papadopoulos2019detecting}} &2019 &\publicdata &\acc &\thirdgen &\chk &\distributionbar{31.11}{68.89} &\text{50Hz} \\
        \midrule
        \multirow{1}{*}{Shawen et al. \cite{shawen2020role}} &2020 &\privatedata &\acc/\gyr &\secondgen &\xmark &\distributionbar{0}{100} &\text{50Hz} \\
        \multirow{1}{*}{Duque et al. \cite{duque2020angular}} &2020 &\privatedata &\gyr &\secondgen &\chk &\distributionbar{23.53}{76.47} &\text{100Hz} \\
        \multirow{1}{*}{De Araujo et al. \cite{de2020hand}} &2020 &\privatedata &\acc/\gyr &\secondgen &\chk &\distributionbar{64}{36} &\text{100Hz} \\
        \multirow{1}{*}{Ranjan et al. \cite{ranjan2021convolutional}} &2020 &\privatedata &\acc &\thirdgen &\xmark &\distributionbar{0}{100} &\text{64Hz} \\
        \multirow{1}{*}{Zhang et al. \cite{zhang2020comparing}} &2020 &\publicdata &\acc &\secondgen &\notr &\distributionbar{17.14}{82.86} &\text{100Hz} \\
        \multirow{1}{*}{Varghese et al. \cite{varghese2020smart}} &2020 &\privatedata &\acc/\gyr/\mymagnetometericon &\secondgen &\chk &\distributionbar{16.04}{83.96} &\text{100Hz} \\
        \multirow{1}{*}{San Segundo et al. \cite{san2020parkinson}} &2020 &\privatedata &\acc &\thirdgen &\xmark &\distributionbar{0}{100} &\text{100Hz} \\
        \multirow{1}{*}{Wu et al. \cite{wu2020assessment}} &2020 &\privatedata &\acc &\secondgen &\xmark &\distributionbar{0}{100} &\text{100Hz} \\
        \multirow{1}{*}{Ibrahim et al. \cite{ibrahim2020parkinson}} &2020 &\privatedata &\notr &\thirdgen &\xmark &\distributionbar{0}{100} &\text{100Hz} \\
        \multirow{1}{*}{Sein Mousavi et al. \cite{sein2020diagnosis}} &2020 &\privatedata &\acc &\secondgen &\notr &\notr &\text{100Hz} \\
        \midrule
        \multirow{1}{*}{Sigcha et al. \cite{sigcha2021automatic}} &2021 &\privatedata &\acc &\thirdgen &\chk &\distributionbar{22.22}{77.78} &\text{50Hz} \\
        \multirow{1}{*}{Aljihmani et al. \cite{aljihmani2020detection}} &2021 &\privatedata &\acc &\secondgen &\notr &\notr &\text{45Hz} \\
        \multirow{1}{*}{Ehsan et al. \cite{ehsan2021automated}} &2021 &\privatedata &\acc &\secondgen &\chk &\distributionbar{41.18}{58.82} &\text{100Hz} \\
        \multirow{1}{*}{Channa et al. \cite{channa2021wear}} &2021 &\privatedata &\gyr &\secondgen &\chk &\distributionbar{50}{50} &\notr \\
        \multirow{1}{*}{Tong et al. \cite{tong2021cnn}} &2021 &\privatedata &\acc/\gyr/\mymagnetometericon &\thirdgen &\chk &\distributionbar{50}{50} &\text{200Hz} \\
        \multirow{1}{*}{Peres et al. \cite{peres2021discrimination}} &2021 &\privatedata &\acc/\gyr/\mymagnetometericon &\secondgen &\chk &\distributionbar{44.44}{55.56} &\text{50Hz} \\
        \multirow{1}{*}{AlMahadin et al. \cite{almahadin2021parkinson}} &2021 &\publicdata &\acc &\secondgen &\xmark &\distributionbar{0}{100} &\text{100Hz} \\
        \multirow{1}{*}{Babayan et al. \cite{babayan2021everyday}} &2021 &\privatedata &\acc &\secondgen &\notr &\notr &\text{50Hz} \\
        \multirow{1}{*}{Legaria Santiago et al. \cite{legaria2022computer}} &2021 &\privatedata &\notr &\secondgen &\chk &\distributionbar{17.14}{82.86} &\notr \\
        \multirow{1}{*}{AlMahadin et al. \cite{almahadin2021task}} &2021 &\publicdata &\acc &\secondgen &\xmark &\distributionbar{0}{100} &\text{100Hz} \\
        \midrule
        \multirow{1}{*}{Sun et al. \cite{sun2023parkinson}} &2022 &\privatedata &\acc/\gyr &\thirdgen &\xmark &\distributionbar{0}{100} &\text{100Hz} \\
        \multirow{1}{*}{Hathaliya et al. \cite{hathaliya2022deep}} &2022 &\publicdata &\acc &\thirdgen &\xmark &\distributionbar{0}{100} &\text{100Hz} \\
        \multirow{1}{*}{Channa et al. \cite{channa2022parkinson}} &2022 &\publicdata &\acc &\thirdgen &\notr &\notr &\text{100Hz} \\
        \midrule
        \multirow{1}{*}{Li et al. \cite{li2023improved}} &2023 &\publicdata &\acc &\secondgen &\xmark &\distributionbar{0}{100} &\text{50Hz} \\
        \multirow{1}{*}{Igene et al. \cite{igene2023machine}} &2023 &\publicdata &\acc &\secondgen &\chk &\distributionbar{50}{50} &\text{31.25Hz} \\
        \multirow{1}{*}{Zhao et al. \cite{zhao2023early}} &2023 &\privatedata &\acc &\secondgen &\chk &\distributionbar{28.13}{71.88} &\text{100Hz} \\
        \multirow{1}{*}{Sun et al. \cite{sun2021tremorsense}} &2023 &\privatedata &\acc/\gyr &\secondgen &\xmark &\distributionbar{0}{100} &\text{100Hz} \\
        \bottomrule
    \end{tabular}
\end{adjustbox}

\begin{adjustbox}{margin=0 0 0 0.1cm, width=0.98\linewidth}
    \begin{tabular}{c@{  }l c@{  }l c@{  }l c@{  }l c@{  }l c@{  }l}
        \publicdata &Public Data &\privatedata &Private Data &\acc &Accelerometer sensor &\gyr &Gyroscope sensor &\mymagnetometericon &Magnetometer sensor &\firstgen &First Generation Model\\
        \secondgen &Second Generation Model &\thirdgen &Third Generation Model &\notr &Not Reported 
    \end{tabular}
\end{adjustbox}

}
	\end{center}
\end{table*}

With this in mind, we will analyze the characteristics according to the proposed taxonomy.

\subsection{Signal Adquisition}

The first characteristic to be analyzed is signal acquisition, a highly significant step. Given the research purpose and the development of guidelines for future applications, achieving the highest possible quality is expected. This is crucial for ensuring the necessary quality in important tasks, such as pathology monitoring and alternative diagnostics \cite{penzel2001acquisition}. Therefore, the analysis will begin with the general composition of the dataset, along with more specific aspects like the type of sensor used and the sampling frequency applied.

One of the initial characteristics to analyze within the \textbf{dataset composition} is \textbf{variability}, explicitly concerning whether studies include a control group. In medical research, a control group is defined as a group that either receives no treatment or receives or receives a standard treatment. Beyond treatment, the concept extends to a group without a pathology diagnosis \cite{harvard_2011}.

\begin{figure}[ht] 
\centering 
\includegraphics[width=0.7\columnwidth]{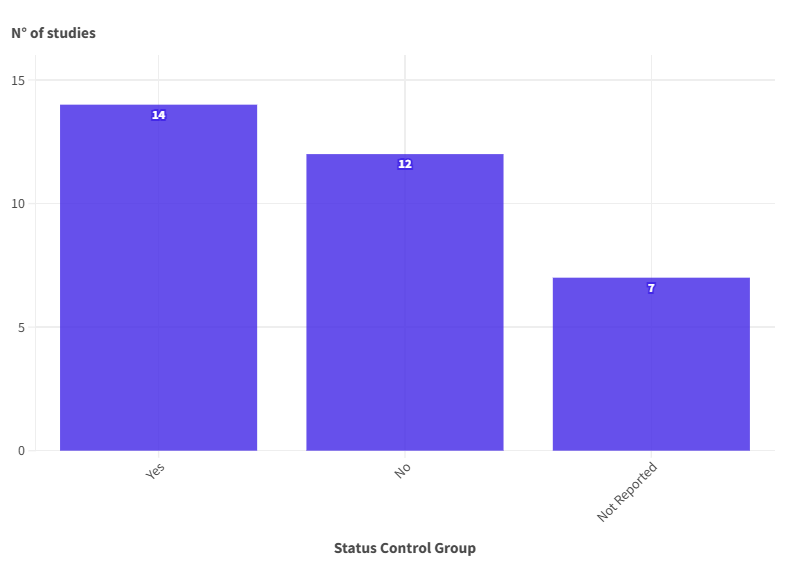}
\caption{Distribution of studies with a control group.}
\label{fig:figure5}
\end{figure}

In the specific context of Figure \ref{fig:figure5}, we observe only 14 studies considering a distribution with a control group. Similarly, Table \ref{tab:table_all} details the specific distributions for each case, which will be a crucial factor in developing the models. Generally, this characteristic indicates that the dataset comprises signals from individuals with and without Parkinson's, though not necessarily in equal proportions.

This inherent variability enables the model to distinguish between ``tremor'' and ``non-tremor,'' serving as an identifier for individuals with Parkinson's. This distinction is essential because tremor recognition in the context of Parkinson's serves two purposes: first, in separating individuals with and without Parkinson's, and second, to sub-classify the severity of tremors in those with Parkinson's. These two aspects are closely tied to the study's specific objectives, making it relevant to discuss key references in this area. Consequently, we will begin by examining studies that do not include a control group.

Ranjan et al. (2021) \cite{ranjan2021convolutional} conducted a study on classifying tremor severity on a scale of 1 to 4, using a dataset of 14 patients with Parkinson's disease and 13 with essential tremors. Although both conditions involve tremors, they exhibit physiological differences, as detailed in Table \ref{tab:Table1} from Saifee et al.'s study (2019) \cite{saifee2019tremor}. The algorithm proposed in the study can generally detect tremors but struggles to distinguish between the two specific conditions and fails to separate tremors from healthy individuals. Another important finding is the characterization of tremors in the 4 to 12 Hz frequency range, which mostly includes essential tremor and certain Parkinson's tremor subtypes (resting, postural, and kinetic), as noted by Raethjen et al. (2009) \cite{raethjen2009cortical} and Deuschl et al. (2000) \cite{deuschl2000pathophysiology}. This frequency range may limit the generality of tremor recognition, as resting tremor, typically within 3 to 6 Hz, falls outside this range, potentially affecting the study's results.

A second study by Wu et al. (2020) \cite{wu2020assessment}, while lacking a control group in the dataset, explores subtyping to distinguish between two Parkinsonian tremors (resting and postural). However, the study does not provide specific considerations for differential characterization. Although the results are positive, the absence of a control group raises the importance of recognizing tremor type or severity and ensuring differentiation from non-pathological signals. Considering that the dataset comprises only patients diagnosed with early-stage Parkinson's, the study presents a model identifying resting and postural tremors. However, limitations arise as it fails to address tremor severity or establish associations with Parkinson's. It can differentiate tremors associated with specific movements but lacks broader diagnostic implications.

Hathaliya et al. (2022) \cite{hathaliya2022deep} focus on distinguishing between Parkinson's disease and essential tremor. Their binary classification exhibits moderately high accuracy, yet the employed characterization for tremor differentiation based on the signal is not reported. This study shares the common challenge of lacking specificity in discriminating between individuals without the pathology associated with Parkinson's and/or essential tremor.

Li et al. (2023) \cite{li2023improved} conducted their study with a dataset with 16 patients diagnosed with Parkinson's, proposing identifying tremor grades ranging from 0 to 4. An important point not addressed is the existence of a grade 0 tremor in patients. This motor characteristic may need clarification regarding how much this feature has been measured. The absence of tremor at grade zero could suggest significance within the diagnosis.
Furthermore, annotations are derived from 12 specific tasks within a 30-second protocol per activity, with 16 annotations per activity. It can be assumed that these were labeled for particular periods, identifying the tremor grade in those intervals. This approach leads to a subdivision that is not centered on individual patients. Similar details regarding Parkinson's patients and the identification of tremor grades from 0 to 4 are found in the work by AlMahadin et al. (2021) \cite{almahadin2021parkinson}. The key distinction lies in the dataset size of 30 patients. While more metrics and processing details are reported, specifics related to the existence of tremor grades are not explicitly outlined.

The details regarding tremor grades for patients diagnosed with Parkinson's are absent in the study presented by Kim et al. (2018) \cite{kim2018wrist}, where labels for tremor grades range from 1 to 4. All patients exhibit motor symptoms, although the utilized protocol has considerable variations. If considering the mentioned protocol, it deviates from the UPDRS protocol and evaluates tremors through a 60-second task in a seated position, corresponding to the resting tremor assessment.

We find a similar approach in the study by San-Segundo et al. (2020) \cite{san2020parkinson}. With 12 patients diagnosed with Parkinson's, the recognition of tremor percentage for daily activities is executed based on the UPDRS scale to measure the tremor grade at a certain point in the protocol. At this stage, when quantifying the percentage of tremor, the metric of tremor grade is lost, reducing it to identifying periods of tremor presence and non-tremor periods. This may provide an idea of binary differentiation and, in a way, loss of information about the tremor grade, making the task less nuanced.

Ibrahim et al. (2020) \cite{ibrahim2020parkinson} have 13 patients diagnosed with Parkinson's for binary classification of \textit{``tremor''} and \textit{``no tremor''}. However, this aspect has been discussed earlier, specifically regarding the existence of patients without tremor presence, which, to some extent, should be observed within evaluations. Considering this feature as part of motor symptoms and the pathology's development, it is expected to be a distinguishing factor in the recognition process. This binary classification approach is present in two more studies, one conducted by Shawen et al. (2020) \cite{shawen2020role} and the other by Sun et al. (2022) \cite{sun2021tremorsense}. The key difference lies in these studies working with 20 and 30 patients, respectively, under the same dynamic of patients diagnosed with Parkinson's and utilizing the UPDRS protocol. This underscores a starting point to consider in future studies encompassing the signal acquisition stage or working with a dataset requiring observations.

In the studies by AlMahadin et al. (2021) and Sun et al. (2023) discussed in \cite{almahadin2021parkinson} and \cite{sun2021tremorsense}, respectively, there are 30 patients, all diagnosed with Parkinson's, for binary classification identifying tremors and non-tremors. In both cases, the UPDRS scale was used as a reference protocol. Still, there is no commentary on how the absence of tremors has been considered and whether these instances can serve as a reference regarding the characterization of the pathology or its degree. Similarly, AlMahadin et al. (2021) present groups differentiated by medication as part of their experiments. One group stopped taking medication during the signal capture, while the other continued. This introduces variability in motor symptomatology between the groups, and the study explores whether the proposed algorithm can generalize effectively to both. The results indicate that the algorithm is not universally applicable, as tremor recognition proved more effective for the group that maintained medication, while precision significantly decreased for the group that ceased medication.

Up to this point, the feature regarding the existence of the control group has been based on the notion that applications can achieve separability between pathological and non-pathological patients, establishing a standard reference for characterizing a non-pathological signal. In this way, the influence of the pathology on a person's movement patterns can be measured, providing insight into motor symptomatology and enabling more specific separability. This plays a crucial role in objectives related to measuring the progression of the pathology and/or personalization among different patient groups.

Considering the control group in studies, seven studies do not provide information on the structure of their dataset or whether they include a control group. We observed this in the studies conducted by Zhang et al. (2020), Aljihmani et al. (2020), Sein Mousavi et al. (2020), Babayan et al. (2021), Thorp et al. (2018), Papadopoulos et al. (2019), and Channa et al. (2022) reported in \cite{zhang2020comparing, aljihmani2020detection, sein2020diagnosis, babayan2021everyday, thorp2018monitoring, papadopoulos2019detecting}, and \cite{channa2022parkinson} respectively. This characteristic leaves a gap regarding whether the results in these studies can differentiate the presence or absence of tremors in Parkinson's pathology and whether they can achieve separability or disease monitoring in the clinical context consistent with the logic of the results. Not having a control group, as defined, prevents differentiation between individuals with pathology and those without. In other words, these applications apply only to a set of pathological individuals, with objectives directed toward that specific target audience.

\begin{figure}[ht] 
\centering 
\includegraphics[width=0.7\columnwidth]{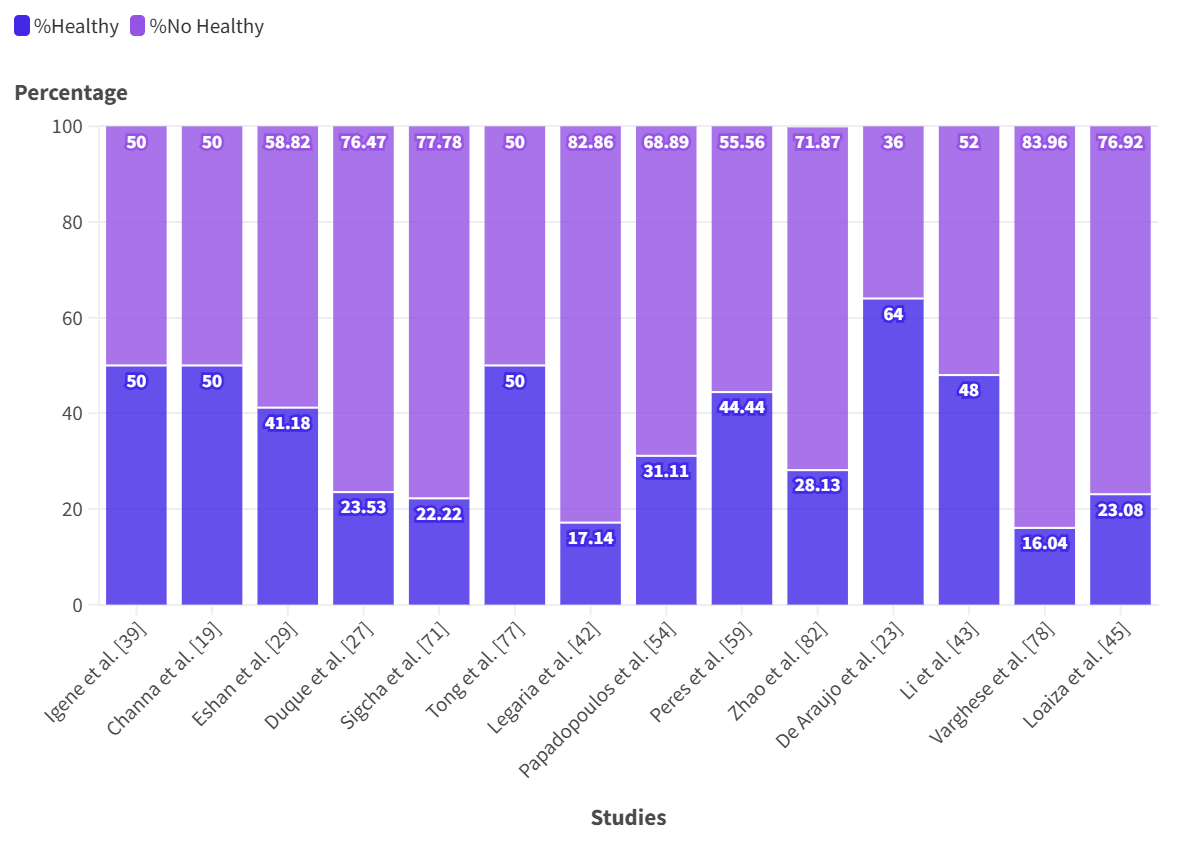}
\caption{Ratio between the control group and pathological group.}
\label{fig:figure6}
\end{figure}

Focusing on having a control group, 14 studies in total include this characteristic. However, this does not necessarily mean that their results are more significant. Instead, it reflects a different focus than studies that lack a control group, depending on each study's approach.  Figure \ref{fig:figure6} illustrates the proportion of control and pathological groups in the evaluated studies. Since machine learning models can be influenced by the data used for training, class imbalance may introduce bias toward the majority class. While an equal class distribution does not automatically make the results more representative, it does highlight ongoing challenges in AI-related to bias introduced by data. Although this bias can be mitigated, it points to the need for a more thorough preprocessing stage, with a preference for balanced distributions to achieve more reliable results for both classes.

There are 14 studies with this characteristic, and three of them have a 50\% distribution between the respective groups. These studies were conducted by Channa et al., Igene et al., and Tong et al. in \cite{igene2023machine, channa2021wear}, and \cite{tong2021cnn}, respectively. This initial reference is important because an equal distribution between classes allows the model to have the same number of observations for each group, theoretically enabling balanced pattern recognition. However, it is worth noting that in Tong et al.'s study, there are only five patients per group, representing a small sample size and limiting the ability to draw conclusive results. This limitation should be taken into account when interpreting the study’s findings. While other studies may not have an equal distribution, this does not invalidate their results. At this stage, data distribution should be viewed as a reference point, with its impact becoming clearer in later stages of analysis. Therefore, the initial class distribution remains a factor to consider but not a definitive measure of a study’s validity.

There are also three studies whose distribution is close to a balanced equilibrium. Ehsan et al. (2021), Peres et al. (2021), and Li et al. (2019) in \cite{ehsan2021automated, peres2021discrimination} and \cite{li2023improved} respectively, have distributions of \textit{41.18\% - 58.82\%}, \textit{44.44\% - 55.56\%}, and \textit{48\% - 52\%}. An important consideration is that only \cite{ehsan2021automated} has a small dataset composed of just 17 individuals. While \cite{peres2021discrimination} and \cite{li2023improved} have 27 and 25 individuals, which, although not significantly different, represent more substantial datasets. The remaining studies, conducted by Duque et al. (2020), Sigcha et al. (2021), Legaria-Santiago et al. (2022), Papadopoulos et al. (2019), Zhao et al. (2023), De Araujo et al. (2020), Varghese et al. (2020), and Loaiza et al. (2019) cited in \cite{duque2020angular, sigcha2021automatic, legaria2022computer, papadopoulos2019detecting, zhao2023early, de2020hand, varghese2020smart} and \cite{loaiza2019using} respectively, show distributions further from equal distribution than the three mentioned at the beginning of the paragraph. 

\begin{figure}[ht] 
\centering 
\includegraphics[width=0.7\columnwidth]{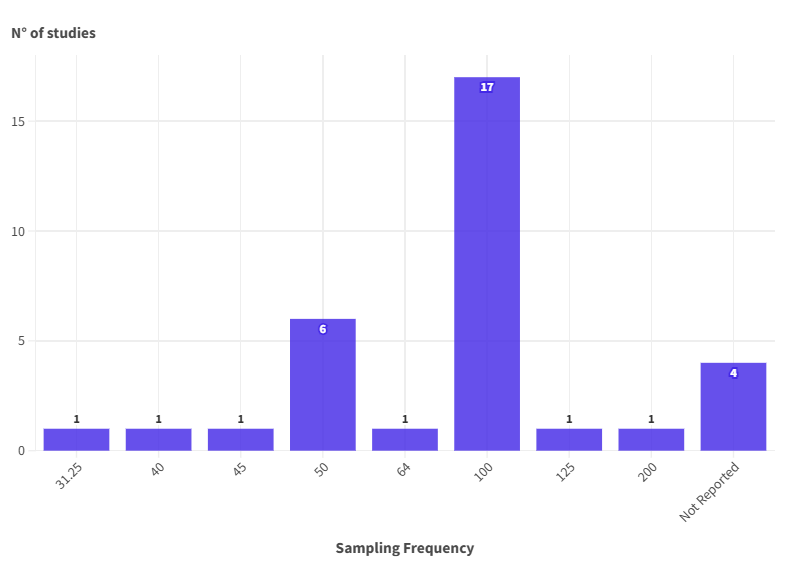}
\caption{Studies with specific frequency.}
\label{fig:figure7}
\end{figure}

Another crucial feature within data acquisition is \textbf{the sampling frequency during signal capture}. The sampling rate provides resolution in the reconstructed signal to capture the specific movement characteristics, such as tremors in specific body parts. Under the concept of effectively capturing signal characteristics, Andreas et al. (2015) in \cite{tobola2015sampling} suggest sampling the signal with at least twice the natural frequency of the movement. Considering that tremors for Parkinson's patients have a maximum frequency of \textit{12 Hz}, a minimum sampling frequency would be \textbf{24 Hz}. Observing Figure \ref{fig:figure7} reveals that all considered studies meet the mentioned requirement. However, it should be clarified that while working with a sampling frequency twice the natural frequency is a minimum limit, the survey by Andreas et al. (2015) concludes that working with a factor of 8 is most advisable. For PD tremor identification, this translates to a frequency of \textbf{96 Hz}, a requirement met by only 19 studies.

\begin{figure}[ht] 
\centering 
\includegraphics[width=0.7\columnwidth]{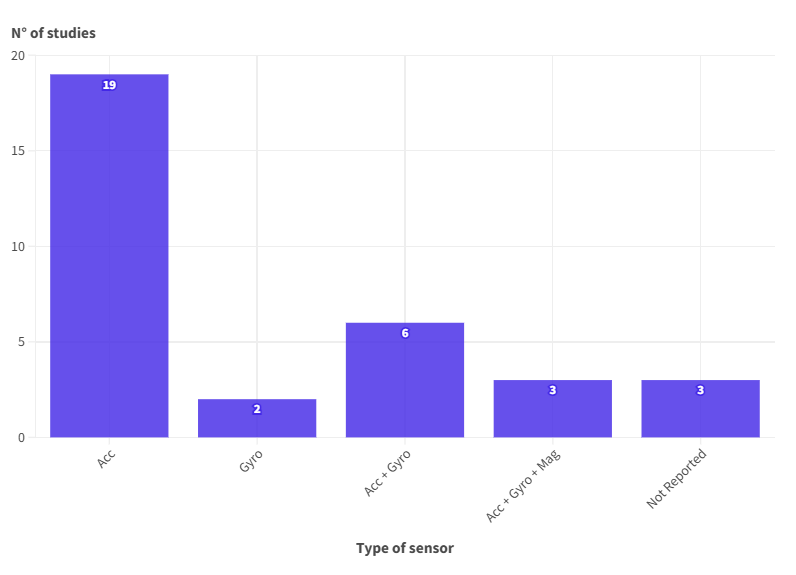}
\caption{Studies with specific sensor.}
\label{fig:figure8}
\end{figure}

Another important characteristic of the studies is the \textbf{type of inertial sensor used for capturing signals}. This feature determines how movement is measured, with the capture quality influenced by the physical principles of each sensor, ultimately affecting signal accuracy. Ilham et al. (2020) in \cite{faisal2019review} define key concepts related to these sensors: accelerometers measure changes in linear velocity and movement direction; gyroscopes measure angular velocity or rotation rate, distinguishing rotational movement from the linear motion captured by accelerometers; and magnetometers measure the intensity and/or direction of the magnetic field around the sensor, indirectly detecting object inclination and movement changes.
Based on these definitions, Figure \ref{fig:figure8} shows that 19 studies used accelerometers as the capture sensor, two used gyroscopes, six utilized both accelerometers and gyroscopes, and three incorporated accelerometers, gyroscopes, and magnetometers. On the other hand, Legaria-Santiago et al. (2022), Papadopoulos et al. (2019), and Ibrahim et al. (2020) in \cite{legaria2022computer, papadopoulos2019detecting, ibrahim2020parkinson} did not report the type of sensor used for signal capture. While their studies may offer hypotheses about the type of signals, these assumptions are not definitive, and the reliability of the results cannot be fully attributed to this characteristic.

The significance of the previous paragraph is that the more accurately the objective movement being measured is characterized, the more it will impact the results and subsequent stages of the study. In this case, the focus is on evaluating tremors in Parkinson's disease, where specific daily life movements are key to assessing tremor severity. Therefore, the ability to accurately capture these signals is critical for later stages of analysis. Most studies follow a standard reference protocol to characterize linear and rotational movements, suggesting that a comprehensive movement assessment should involve inertial sensors such as accelerometers and gyroscopes. The magnetometer could provide valuable insights, especially for movements involving rotations, further enhancing the overall characterization of the tremor.

To conclude the analysis section of the initial signal acquisition stage, we emphasize several key features. First, the inclusion of a control group is crucial for distinguishing between pathological and non-pathological signals, as well as for identifying distinct classes within both groups. When examining the dataset structure, it is important to assess class distribution, as this provides a reference for detecting potential biases in the algorithms developed in these studies. Next, the focus shifts to the intrinsic characteristics of signal capture, such as sampling frequency and the type of sensor used. These factors are critical for evaluating the system's ability to accurately characterize the measured movements, ensuring alignment with the reference protocols for assessing Parkinson’s disease pathology.

\subsection{Data Preprocessing}

In this second stage, we will analyze features related to the signal preprocessing phase, where the signal is prepared for subsequent stages. This includes signal filtering, transformations, feature extraction, and alternative conversions like normalization or segmentation.

A key aspect of signal conditioning is removing irrelevant information for the task at hand, which is closely linked to our case's characterization of the pathology. Since tremor characterization relies on specific frequency bands, the first feature we will examine is the \textbf{frequency bands} used in each study.

\begin{figure}
    \centering
    \resizebox{!}{0.7\textheight}{%
        \tikzset{
    event/.style={
        circle,
        fill,
        inner sep=1pt,
    },
    base/.style={
        draw opacity=1.0,
        fill opacity=1.0,
        text opacity=1.0,
    },
    bar/.style={
        base,
        fill opacity=0.7,
    },
    marker label/.style 2 args={
        postaction={
            decorate,
            decoration={
                markings,
                mark=at position 0.5 with {\node[text=black, rectangle, #1, inner sep=2pt] {#2};},
            },
        },
    },
    left marker/.style={
        rotate=0,
        marker label={left}{#1},
    },
    right marker/.style={
        rotate=0,
        marker label={center}{#1},
    },
    every node/.style={
        font=\small,
    },
}

\tikzexternaldisable
\begin{tikzpicture}

\node[anchor=east] at (0,0.0) {Ranjan et al. \cite{ranjan2021convolutional}};
\node[anchor=east] at (0,-0.6) {Igene et al. \cite{igene2023machine}};
\node[anchor=east] at (0,-1.2) {Channa et al. \cite{channa2021wear}};
\node[anchor=east] at (0,-1.8) {Ehsan et al. \cite{ehsan2021automated}};
\node[anchor=east] at (0,-2.4) {Duque et al. \cite{duque2020angular}};
\node[anchor=east] at (0,-3.0) {Wu et al. \cite{wu2020assessment}};
\node[anchor=east] at (0,-3.6) {Sigcha et al. \cite{sigcha2021automatic}};
\node[anchor=east] at (0,-4.2) {Tong et al. \cite{tong2021cnn}};
\node[anchor=east] at (0,-4.8) {Zhang et al. \cite{zhang2020comparing}};
\node[anchor=east] at (0,-5.4) {Legaria-Santiago et al. \cite{legaria2022computer}};
\node[anchor=east] at (0,-6.0) {Hathaliya et al. \cite{hathaliya2022deep}};
\node[anchor=east] at (0,-6.6) {Papadopoulos et al. \cite{papadopoulos2019detecting}};
\node[anchor=east] at (0,-7.2) {Aljihmani et al. \cite{aljihmani2020detection}};
\node[anchor=east] at (0,-7.8) {Sein Mousavi et al. \cite{sein2020diagnosis}};
\node[anchor=east] at (0,-8.4) {Peres et al. \cite{peres2021discrimination}};
\node[anchor=east] at (0,-9.0) {Zhao et al. \cite{zhao2023early}};
\node[anchor=east] at (0,-9.6) {Babayan et al. \cite{babayan2021everyday}};
\node[anchor=east] at (0,-10.2) {De Araujo et al. \cite{de2020hand}};
\node[anchor=east] at (0,-10.8) {Li et al. \cite{li2019monitoring}};
\node[anchor=east] at (0,-11.4) {Thorp et al. \cite{thorp2018monitoring}};
\node[anchor=east] at (0,-12.0) {Li et al. \cite{li2023improved}};
\node[anchor=east] at (0,-12.6) {Papadopoulos et al. \cite{papadopoulos2019multiple}};
\node[anchor=east] at (0,-13.2) {Channa et al. \cite{channa2022parkinson}};
\node[anchor=east] at (0,-13.8) {San Segundo et al. \cite{san2020parkinson}};
\node[anchor=east] at (0,-14.4) {Ibrahim et al. \cite{ibrahim2020parkinson}};
\node[anchor=east] at (0,-15.0) {AlMahadin et al. \cite{almahadin2021parkinson}};
\node[anchor=east] at (0,-15.6) {Sun et al. \cite{sun2021tremorsense}};
\node[anchor=east] at (0,-16.2) {Shawen et al. \cite{shawen2020role}};
\node[anchor=east] at (0,-16.8) {AlMahadin et al. \cite{almahadin2021task}};
\node[anchor=east] at (0,-17.4) {Varghese et al. \cite{varghese2020smart}};
\node[anchor=east] at (0,-18.0) {Sun et al. \cite{sun2023parkinson}};
\node[anchor=east] at (0,-18.6) {Loaiza et al. \cite{loaiza2019using}};
\node[anchor=east] at (0,-19.2) {Kim et al. \cite{kim2018wrist}};

\node[anchor=east] at (0,-19.8) {Teoric Rest Tremor \cite{saifee2019tremor}};
\node[anchor=east] at (0,-20.4) {Teoric Postural Tremor \cite{saifee2019tremor}};
\node[anchor=east] at (0,-21.0) {Teoric Kinetic Tremor \cite{saifee2019tremor}};
\node[anchor=east] at (0,-21.6) {Teoric Drugs Tremor \cite{saifee2019tremor}};

\node[anchor=east] at (20.7,-22.9) {Hz};

\draw[bar, fill=blue] (4,0.2) rectangle (12,-0.2);

\draw[bar, fill=red] (0,-0.4) rectangle (20,-0.8);

\draw[bar, fill=blue] (4,-1.0) rectangle (12,-1.4);

\draw[bar, fill=green] (3,-1.6) rectangle (6,-2.0);

\draw[bar, fill=blue] (1,-2.2) rectangle (8,-2.6);

\draw[bar, fill=red] (0,-2.8) rectangle (20,-3.2);

\draw[bar, fill=green] (3.5,-3.4) rectangle (7.5,-3.8);

\draw[bar, fill=blue] (3,-4.0) rectangle (6,-4.4);

\draw[bar, fill=blue] (1,-4.6) rectangle (12,-5.0);

\draw[bar, fill=blue] (2,-5.2) rectangle (18,-5.6);

\draw[bar, fill=red] (0,-5.8) rectangle (20,-6.2);

\draw[bar, fill=blue] (1,-6.4) rectangle (12,-6.8);

\draw[bar, fill=green] (3,-7.0) rectangle (12,-7.4);

\draw[bar, fill=red] (0,-7.6) rectangle (20,-8.0);

\draw[bar, fill=green] (1,-8.2) rectangle (16,-8.6);

\draw[bar, fill=blue] (4,-8.8) rectangle (7,-9.2);

\draw[bar, fill=blue] (1,-9.4) rectangle (13,-9.8);

\draw[bar, fill=blue] (1,-10.0) rectangle (20,-10.4);

\draw[bar, fill=blue] (3,-10.6) rectangle (8,-11.0);

\draw[bar, fill=blue] (3.5,-11.2) rectangle (7.5,-11.6);

\draw[bar, fill=blue] (4,-11.8) rectangle (13,-12.2);

\draw[bar, fill=blue] (1,-12.4) rectangle (12,-12.8);

\draw[bar, fill=blue] (3,-13.0) rectangle (12,-13.4);

\draw[bar, fill=blue] (0.3,-13.6) rectangle (12,-14.0);

\draw[bar, fill=blue] (3,-14.2) rectangle (17,-14.6);

\draw[bar, fill=green] (3,-14.8) rectangle (6,-15.2);
\draw[bar, fill=brown] (6,-14.8) rectangle (9,-15.2);
\draw[bar, fill=lightbrown] (9,-14.8) rectangle (12,-15.2);

\draw[bar, fill=blue] (2,-15.4) rectangle (10,-15.8);

\draw[bar, fill=red] (1,-16.0) rectangle (20,-16.4);

\draw[bar, fill=green] (3,-16.6) rectangle (6,-17.0);
\draw[bar, fill=brown] (6,-16.6) rectangle (9,-17.0);
\draw[bar, fill=lightbrown] (9,-16.6) rectangle (12,-17.0);

\draw[bar, fill=red] (0,-17.2) rectangle (20,-17.6);

\draw[bar, fill=red] (0,-17.8) rectangle (20,-18.2);

\draw[bar, fill=green] (4,-18.4) rectangle (6,-18.8);
\draw[bar, fill=brown] (6,-18.4) rectangle (8,-18.8);

\draw[bar, fill=blue] (1,-19.0) rectangle (20,-19.4);

\draw[black, thick] (-3.6,-19.5) -- (21,-19.5);
\draw[bar, fill=green] (3,-19.6) rectangle (7,-20.0);
\draw[bar, fill=brown] (7,-20.2) rectangle (12,-20.6);
\draw[bar, fill=lightbrown] (4,-20.8) rectangle (9,-21.2);
\draw[bar, fill=orange] (8,-21.4) rectangle (12,-21.8);


\foreach \x in {0,1,...,20}
    \draw (\x,-22.2) -- (\x,-22.7) node[below] {\x};

\begin{scope}[shift={(0,-24.2)}]
    \draw[fill=lightbrown] (0,0) rectangle (1,0.5); \node[right] at (1.2,0.25) {Kinetic Tremor};
    \draw[fill=green] (3.5,0) rectangle (4.5,0.5); \node[right] at (4.7,0.25) {Rest Tremor};
    \draw[fill=brown] (7,0) rectangle (8,0.5); \node[right] at (8.2,0.25) {Postural Tremor};
    \draw[fill=orange] (10.8,0) rectangle (11.8,0.5); \node[right] at (12.0,0.25) {Drugs};
    \draw[fill=red] (13.2,0) rectangle (14.2,0.5); \node[right] at (14.4,0.25) {Not Reported};
    \draw[fill=blue] (16.8,0) rectangle (17.8,0.5); \node[right] at (18.0,0.25) {All Tremor};
\end{scope}

\end{tikzpicture}

\tikzexternalenable
        }
    \caption{Overview of the frequency bands used across various studies for tremor analysis, categorized by the type of tremor—kinetic, rest, postural, and drug-induced. The horizontal axis represents the frequency range in Hz, while the vertical axis lists the studies by their first author. Color-coded bars indicate the frequency ranges reported or not reported, with green representing kinetic tremor, brown for rest tremor, light brown for postural tremor, orange for drug-related studies, red for studies where the frequency was not reported, and blue for studies encompassing all tremor types. The bottom section shows theoretical frequency ranges for comparison. This figure highlights the variability in frequency band usage and the occasional lack of specific reporting across the literature.
    }%
    \label{fig:figure9}
\end{figure}
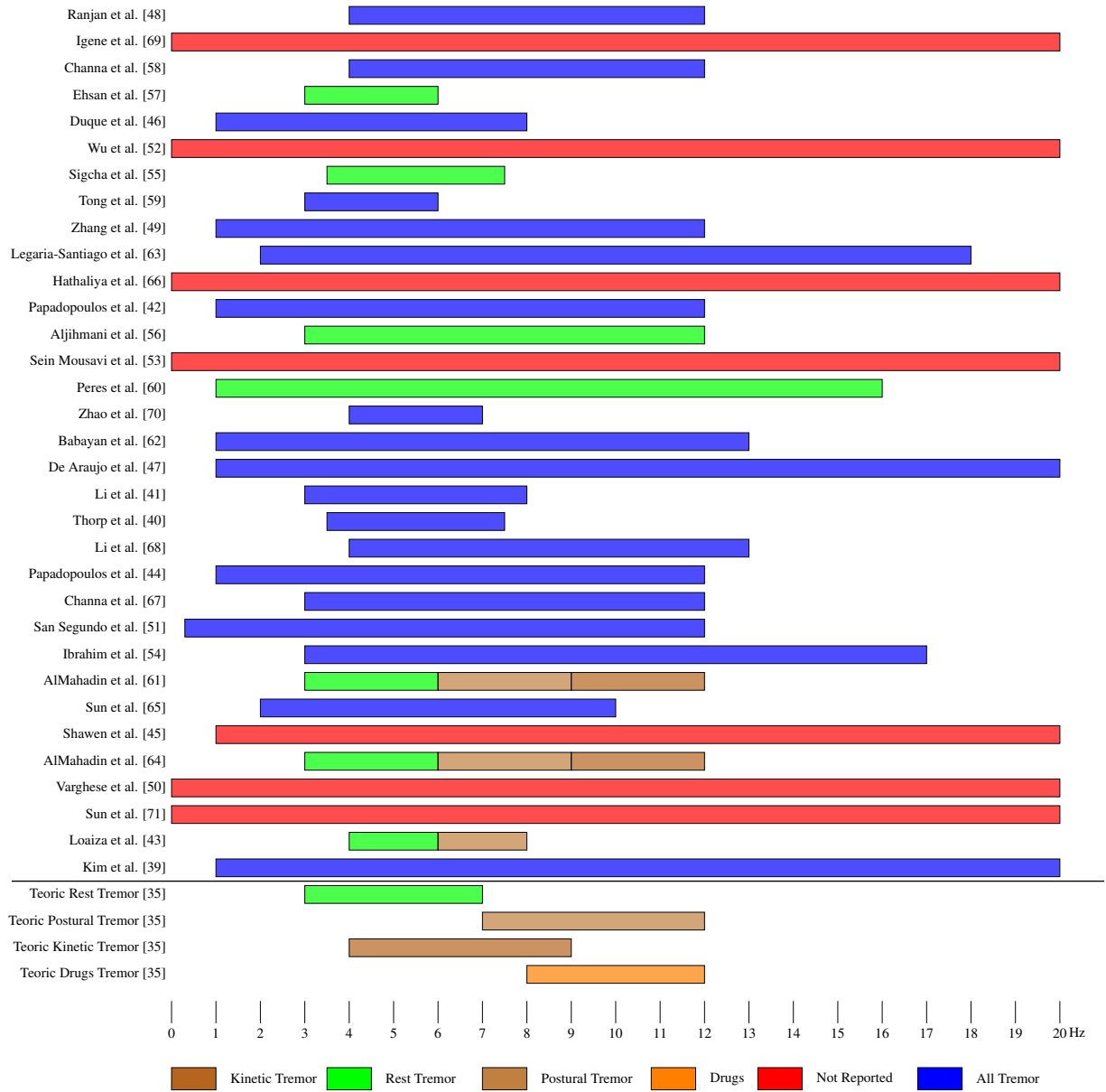

Figure \ref{fig:figure9} summarizes the frequency bands considered across the studies. These bands are reported using physiological references for tremor characterization, as indicated by Raethjen et al. (2009)~\cite{raethjen2009cortical} and Deuschi et al. (2000)~\cite{deuschl2000pathophysiology}, who identify a range of 3 to 12 Hz for tremors at rest, as well as kinetic, postural, and drug-induced tremors.

Regarding Figure \ref{fig:figure9}, seven studies do not specifically mention the consideration of frequency bands, suggesting that they worked with raw signals without applying a preprocessing stage focused on frequency-level filtering. These studies include Igene et al. (2023), Wu et al. (2021), Hathaliya et al. (2021), Sein Mousavi et al. (2022), Shawen et al. (2021), Varghese et al. (2020), and Sun et al. (2022), reported in \cite{igene2023machine, duque2020angular, hathaliya2022deep, sein2020diagnosis, shawen2020role, varghese2020smart} and \cite{sun2021tremorsense}, respectively. While this does not necessarily invalidate their results, this omission is significant because frequency band filtering is crucial for removing irrelevant information in tremor recognition. The absence of such preprocessing raises concerns about potential deficiencies in data treatment. However, it is possible that these gaps were addressed through other forms of processing or by the model itself. However, while additional processing might enhance the outcomes, there remains uncertainty as to whether the performance could have improved further by removing more specific noise from the data.

In line to process a model using specific information that characterizes motor symptoms, the key reference is the physiological data within the 3 to 12 Hz range, which is commonly associated with Parkinson’s pathology, assuming no additional tremor factors are present, as outlined in Table \ref{tab:Table1}. We begin by considering the work of Ranjan et al. (2021) and Channa et al. (2021) in \cite{ranjan2021convolutional} and \cite{channa2021wear}, where the frequency band selected ranges from 4 to 12 Hz. However, it is unclear what rationale underlies this specific frequency choice, especially since their experimental protocols account for all three types of tremors found in Parkinson’s disease. The omission of information below 4 Hz could directly affect the accuracy of their results, as it may exclude relevant tremor data critical to comprehensive symptom analysis.

Ehsan et al. (2021) and Tong et al. (2021), as reported in \cite{ehsan2021automated} and \cite{tong2021cnn}, consider a frequency band ranging from 3 to 6 Hz, which does not fully encompass the complete reference band, nor does it entirely cover the range associated with resting tremor. This limitation is notable because both studies aim to differentiate between Parkinson's patients based on protocols considering resting, kinetic, and postural tremors. Given this, it is possible that important signal characteristics were excluded by applying such a narrow filter. This raises the question of how much relevant information was lost and whether features captured only within the 3 to 6 Hz band are sufficient for effectively distinguishing between Parkinson's patients and non-pathological subjects.

Some studies have found that they do not fully cover the frequency band associated with tremors in Parkinson's disease. For example, Sigcha et al. (2021) in \cite{sigcha2021automatic} reported working with a frequency band of 3.5 to 7.5 Hz, which only partially covers the range associated with resting tremor, even though it was the main focus of their study. Similarly, Zhao et al. (2023) in \cite{zhao2023early} worked with a band of 4 to 7 Hz, which overlaps with the resting tremor band but still does not fully encompass it. Thorp et al. (2018) in \cite{thorp2018monitoring} also encountered this issue, covering a 3.5 to 7.5 Hz band. Loaiza et al. (2019) in \cite{loaiza2019using} worked in the 4 to 8 Hz band, and Sun et al. (2023) in \cite{sun2023parkinson} used a band of 2 to 10 Hz. In all these cases, the frequency bands employed do not fully cover specific tremors—such as resting, postural, or kinetic tremors or combinations thereof. This raises the need to justify using these specific bands when they do not align with the physiological reference bands associated with Parkinson’s tremors.

Conversely, other studies cover the full frequency range of the three main types of tremors (resting, postural, and kinetic), typically from 3 to 12 Hz. For example, Duque et al. (2020) in \cite{duque2020angular} used a band of 1 to 16 Hz; Legaria et al. (2022) in \cite{legaria2022computer} worked within 2 to 18 Hz; Papadopoulos et al. (2019) in \cite{papadopoulos2019detecting} and De Araujo et al. (2020) in \cite{de2020hand} both employed bands from 1 to 12 Hz. Other studies, such as those by Peres et al. (2021) in \cite{peres2021discrimination} and Babayan et al. (2021) in \cite{babayan2021everyday}, used bands from 1 to 16 Hz and 1 to 13 Hz, respectively. San Segundo et al. (2020) in \cite{san2020parkinson} utilized a range of 0.3 to 12 Hz, while Ibrahim et al. (2020) in \cite{ibrahim2020parkinson} reported a working band of 3 to 17 Hz. Finally, Kim et al. (2018) in \cite{kim2018wrist} used a band of 1 to 20 Hz. These broader bands capture physiological tremor characteristics and additional information not typically associated with the pathology. This prompts an exploration into whether this extra information aids in prediction or is simply being filtered out by the models. However, this distinction is not addressed in the studies mentioned.

Next, we discuss studies focusing on the specific frequency band for characterizing tremors based on physiological evidence, defined as 3 to 12 Hz. While this frequency band does not necessarily guarantee the best results across all studies or tasks, it provides a clear framework for including relevant characteristics specific to the pathology when analyzing signal processing sequences. The key studies in this context are Aljihmani et al. (2020), Channa et al. (2022), and AlMahadin et al. (2021), as reported in \cite{aljihmani2020detection}, \cite{channa2022parkinson}, and \cite{almahadin2021task}, respectively. There are significant differences in the results among these studies and compared to other complementary studies. Given this variability, assessing whether certain parts of the frequency band are more important than others and whether relevant information might exist outside the defined band is crucial.

Finally, in the first part of this section, the working frequency bands were analyzed within the context of a preprocessing stage. Despite differences in the results, there is a clear discrepancy in the choice of frequency bands. Some studies cover only part of the physiological reference band, while others encompass the full range, including frequencies that may not be relevant for diagnosing Parkinson's disease. Furthermore, many studies do not clearly justify selecting one band over another, suggesting that these choices were largely experimental. This indicates that the bands cannot be generalized, and a specific range of importance within the full band has yet to be definitively identified.

\subsection{Feature extraction}

This section focuses on the features extracted from signals, which are used as specific representations to theoretically differentiate between pathological and non-pathological signals based on their characteristic features. Initially, we observe features in the temporal, frequency, and statistical domains, noting that many studies combine both temporal and frequency features. In Parkinson's disease, where frequency characteristics are prominent, specific individuals or combinations of frequency features could explicitly differentiate the pathological group from the non-pathological group.

For\textbf{ temporal features}, a set of references appears similar yet variable across studies. To provide a clearer understanding, we discuss these features collectively. They include \textbf{mean}, s\textbf{tandard deviation}, \textbf{variance}, \textbf{root mean square} (RMS), \textbf{peak value}, \textbf{zero-crossing rate} (the number of times the signal crosses zero), \textbf{mean absolute values} (MAV), and \textbf{the first difference of mean absolute values} (MAVFD), among others. Although there may not always be a specific justification for choosing one feature over another, some features, such as mean and standard deviation, are consistently used in nearly all studies. Extracting features aims to identify those with the highest representative capacity.

Similarly, \textbf{frequency features} are often based on \textbf{Power Spectral Density} (PSD), which describes how signal power is distributed in the frequency domain, representing the energy in respective frequency bands. Since the tremors characteristic of Parkinson's disease are closely related to frequency, these features can be particularly relevant. Among the frequency features commonly reported are \textbf{mean PSD}, \textbf{standard deviation of PSD}, \textbf{power band} (BP), \textbf{peak power frequency} (PPF), \textbf{harmonic index} (HI), \textbf{relative energy} (RE), \textbf{harmonic index ratio} (HIR), and \textbf{sum of maximum power} (SMP). These spectral density features help differentiate pathological signals from non-pathological ones.

The final set of features is based on \textbf{statistical concepts} such as \textbf{mode}, \textbf{median}, \textbf{Shannon entropy}, \textbf{skewness}, and \textbf{kurtosis}, applied to both temporal and frequency domains to measure signal dispersion. Features related to the degree of disorder, such as Shannon entropy, are expected to reveal differences due to the erratic behavior in tremors. Thus, a wide range of features is employed for tasks like binary and multilabel classification, although temporal, frequency, and statistical features are not always used together.

For example, Igene et al. (2023) in \cite{igene2023machine} used only temporal and statistical features for binary classification between Parkinson's and non-Parkinson's cases. Similarly, San Segundo et al. (2020) in \cite{san2020parkinson} extracted a different set of temporal and statistical features. While Igene et al. (2023) achieved a 95.0\% AUC, San Segundo et al. (2020) reported an 88.7\% AUC, highlighting notable differences in representation capacity across specific feature sets. However, this comparison is not definitive, as the results depend on the signal characteristics and the task at hand.

As previously mentioned, frequency domain features may offer greater representational power due to their direct relationship with Parkinson’s tremors. The studies by Duque et al. (2020) \cite{duque2020angular}, Sigcha et al. (2021) \cite{sigcha2021automatic}, and Loaiza et al. (2019) \cite{loaiza2019using} considered only frequency features for binary classification. Duque et al. (2020) achieved the highest classification metric, exceeding 90\%, due to the use of a larger number of features. Sigcha et al. (2021) and Loaiza et al. (2019), with fewer features, achieved comparable results, suggesting that signal quality and preprocessing may also influence performance.

No study relied solely on statistical features for classification; instead, temporal and/or statistical features were often combined with frequency features. For example, Channa et al. (2021) \cite{channa2021wear} combined temporal and frequency features, achieving over 83\% accuracy in multilabel classification. Similar results were found in studies by Zhang et al. (2020) \cite{zhang2020comparing}, Legaria et al. (2022) \cite{legaria2022computer}, Peres et al. (2021) \cite{peres2021discrimination}, Zhao et al. (2023) \cite{zhao2023early}, Li et al. (2023) \cite{li2023improved}, San Segundo et al. (2020) \cite{san2020parkinson}, AlMahadin et al. (2021) \cite{almahadin2021parkinson}, and Kim et al. (2018) \cite{kim2018wrist}. Although results vary across studies, some features—such as mean, standard deviation, power band, and peak frequency of the PSD—are consistently used, highlighting their potential representational power.

In addition, some studies utilized a combination of temporal, frequency, and statistical features for classification tasks. For instance, Ehsan et al. (2021) \cite{ehsan2021automated}, Babayan et al. (2021) \cite{babayan2021everyday}, De Araujo et al. (2020) \cite{de2020hand}, Li et al. (2019) \cite{li2019monitoring}, Sun et al. (2023) \cite{sun2023parkinson}, and Shawen et al. (2020) \cite{shawen2020role} achieved different results, indicating that combining various feature types can improve the separation between classes, such as individuals with and without Parkinson's. These studies suggest that frequency features, complemented by temporal and/or statistical features, play a crucial role in differentiating individuals with Parkinson’s. This is particularly evident in De Araujo et al. (2020), which achieved accuracies of 96.4\% (training) and 98.4\% (testing), as well as Li et al. (2019) with 92.0\% accuracy, and Sun et al. (2023) with 92.28\%.

In conclusion, the results of these studies depend heavily on the extracted features and the data quality, particularly regarding the nature of the signals. Therefore, the reported features in these studies may provide a valuable starting point for identifying effective features for classification.

\subsection{Selection and Evaluation of AI models}

In this section, we analyze the development of studies from the models' perspective, following the sub-classification proposed by Skandha et al. (2020) in \cite{skandha20203}, which differentiates three groups of models. As mentioned earlier, 23 studies utilized models from the second generation, encompassing classical machine-learning models inherently linked to feature extraction and/or signal adaptation using signal flattening. The development of the third generation, which includes deep learning models, is represented by 11 studies, while one study does not report on the model used. Specifically, this last study was developed by Thorp et al. (2018) in \cite{thorp2018monitoring}, which, despite reporting sensitivity and specificity results above 99\%, does not comment on the model or the extracted features used to achieve the reported results. In fact, for this study, it is unknown whether control group conditions, model generation, data distribution, and capture frequency are found within the experiment, as seen in Table \ref{tab:table_all}. 

Of the 23 models belonging to the second generation, multiple models were implemented and analyzed in several studies due to their classical nature. Specifically, 12 studies implemented the \textbf{Support Vector Machine} (SVM) model, with varying results but maintaining metrics above 76\%. The highest performance in terms of classification accuracy was reported by AlMahadin et al. (2021) in \cite{almahadin2021parkinson}, with an accuracy of 98\%. In line with the previous subsection, this high performance resulted from a characterization that included temporal, frequency, and statistical features. The study compared classical models to find the best result for the classification task. Feature extraction was performed on 4-second signal segments with a 50\%  overlap, yielding 102 features for each signal. 

The remaining 11 studies that used the SVM model show similar performance, closely mirroring the results by AlMahadin et al. (2021) reported in \cite{almahadin2021parkinson}. For instance, Igene et al. (2023) reported in \cite{igene2023machine} employed a similar methodology with variations in the window size to 2 seconds and normalization of signals, resulting in a higher number of features and an accuracy of 94.4\%. Zhang et al. (2020) reported in \cite{zhang2020comparing} used a 3-second window and fewer features overall, as this study did not use statistical features, achieving an accuracy of 85.9\%. Sein Mousavi et al. (2020) reported in \cite{sein2020diagnosis} did not comment on a window-based methodology, performing feature extraction on the entire signal, achieving an accuracy of 91.35\%. Meanwhile, Peres et al. (2021) reported in \cite{peres2021discrimination} combined three types of sensors, totaling six sensors and 108 temporal and frequency features, achieving an accuracy of 76.90\%. Zhao et al. (2023) reported in \cite{zhao2023early} employed a feature selection methodology based on correlation, achieving an accuracy of 96.15\%. Li et al. (2023) reported in \cite{li2023improved}, used a feature extraction methodology on temporal and frequency features in windows, achieving an accuracy of 97.96\%, which is closest to AlMahadin et al. (2021) reported in \cite{almahadin2021task} and notable for its high metric within all considered studies. The last three studies not yet mentioned, conducted by Li et al. (2019) in \cite{li2019monitoring}, Sun et al. (2023) in \cite{sun2023parkinson}, and Varghese et al. (2020) in \cite{varghese2020smart}, reported accuracies of 92\%, 92.28\%, and 79.00\% respectively. Li et al. (2019) reported in \cite{li2019monitoring} used a window-based methodology differing from those proposed by Sun et al. (2023) in \cite{sun2023parkinson} and Varghese et al. (2020) in \cite{varghese2020smart}. The results consistently show high performance. However, there needs to be a clear consensus on the significance of any particular feature or set of features across these studies. 

Continuing with the review of studies belonging to the second generation, we have six studies that used the \textbf{K-nearest neighbors} or KNN model. Starting with the study by Channa et al. (2021) in \cite{channa2021wear}, which conducted a multiclass classification to identify "healthy," "bradykinesia," and "tremor" classes, achieving 100\% sensitivity and specificity for the "tremor" class with a total of 10 captures in the dataset for this class using frequency and temporal features in the 2-20 Hz band. Duque et al. (2020) in \cite{duque2020angular} reported an accuracy of 77.8\% for distinguishing between Parkinson's patients and Essential Tremor, with a variation of 9.9\% according to the tests performed. Similarly, the sensitivity and specificity below 80\% remain a reference for the features used. Aljihmani et al. (2020) reported in \cite{aljihmani2020detection} achieved an accuracy of 93.70\% for differentiating "rest tremor" from "postural tremor." However, it is worth noting that the dataset comprised 12 individuals, all diagnosed with Parkinson's, and lacking a control group, which may make the differentiation inconclusive. De Araujo et al. (2020) in \cite{de2020hand} reported an accuracy of 96.9\% using temporal and frequency features. Lastly, Loaiza et al. (2019) in \cite{loaiza2019using} reported a differentiation between "Essential Tremor" and "Parkinson's" with a sensitivity of 73.8\% and specificity of 76.4\%, as well as differentiating "Tremor" and "No tremor" with a sensitivity of 73.2\% and specificity of 96.9\%. These results stem from features derived from the power spectral density (PSD), highlighting the importance and weight of frequency features for classification. However, it should be considered that the dataset lacked a control group, meaning the differentiation for recognizing individuals with Parkinson's cannot be generalized, a point previously noted. Lastly, Babayan et al. (2021) in \cite{babayan2021everyday} should have reported metrics on the experiments, comparing the model choice and the inconclusive results. 

Similarly, we found only four studies using the \textbf{Random Forest} (RF) model. We start with Ehsan et al. (2021) reported in \cite{ehsan2021automated}, which focused on recognizing resting tremor, achieving an accuracy of 88.33\%. Wu et al. (2020) reported in \cite{wu2020assessment} tested an RF model among others but found better performance for a neural network, with the RF model accuracy reported below 85\%, but higher than that proposed by AlMahadin et al. (2021) in \cite{almahadin2021task}, which reported results of 80\% and 77\% for two experiments conducted on two datasets differentiated by the presence of medication, showing variable results. The last work related to the RF model was conducted by Shawen et al. (2020), reported in \cite{shawen2020role}, which achieved 79\% accuracy for distinguishing between Parkinson's and non-Parkinson's, showing high metrics despite the task differentiation, reflecting the representation of extracted features. 

Concluding the review of second-generation studies, we have the study by Legaria-Santiago et al. (2022) reported in \cite{legaria2022computer}, which used a \textbf{Fuzzy Logic} model for multiclass identification, achieving accuracies of 89\%, 82\%, 100\%, and 100\% for tremor scores of 0, 1, 2, and 3, respectively. The variations in extracted features indicate the importance of balancing features with the applied model. Although we cannot definitively conclude the greater significance of features over the model, the studies provide good indications. 

Studies related to feature extraction show excellent results for binary and multiclass classification processes, highlighting differences in feature combinations. The differences are seen in the individual use of temporal, frequency, and statistical features compared to a combination of these, where the performance is superior. This aligns with the definition of pathology characterization, where it can be characterized temporally and complemented with frequency information, providing complementary information to the model and ultimately better characterizing the pathology through its motor symptoms. 

Following the model generation framework proposed by Skandha et al. (2020) in \cite{skandha20203}, we discuss the third-generation models. A total of 11 studies related to deep architectures were identified, with the highest accuracy of all studies being 97.32\% for a \textbf{Convolutional Neural Network} (CNN) model developed by Tong L. et al. (2021) in \cite{tong2021cnn}. Although this reference shows excellent results, the remaining studies using deep networks show few differences in results. For example, Ranjan et al. (2021) in \cite{ranjan2021convolutional} used a CNN model to classify the degree of tremor (scores 1, 2, 3, and 4) after transforming the signal into a 2D image, obtaining accuracies of 100\%, 91\%, 75\%, and 66\% for scores 1, 2, 3, and 4, respectively. This is likely consistent with the amount of data for each tremor grade and the prevalence of these grades in a clinical setting. However, the study does not mention this as it uses a private dataset. In the same vein of transforming signals into 2D images, the study by Sun M. et al. (2021) in \cite{sun2021tremorsense} creates a stack of images for feature extraction, achieving an accuracy of 93.31\%, while Kim H. et al. (2018) in \cite{kim2018wrist} uses the same signal-to-image transformation methodology to generate features from these images, achieving an accuracy of 85.00\%, indicating the feasible use of image-based representations for tremor recognition. 

With a different approach to signal usage, Sigcha et al. (2021) in \cite{sigcha2021automatic} conducts experiments based on the use of CNNs for the detection of resting tremor, reporting a sensitivity and specificity of 86.10\% under a ROC curve of 0.936, introducing the concept of "context" for providing a reference within the classifier for tremor recognition. Another approach based on a CNN is found in the work of Papadopoulos et al. (2019) in \cite{papadopoulos2019detecting}, which, through a resampling methodology and windowing method, achieves a result of 95.50\% accuracy with a sensitivity of 82.80\%, considering feature extraction separated by patient, introducing the concept of "subject bag," which consists of aggregating features for each patient individually. This windowing methodology is also developed by San Segundo et al. (2020) in \cite{san2020parkinson}, where they consider a segmentation size of 3 seconds, then perform a Fourier transformation and extract frequency-related features, achieving an AUC of 88.70\% within the developed CNN model. 

Finally, some studies distinguished by the transformation or architecture are discussed below. In the first study by Ibrahim et al. (2020) in \cite{ibrahim2020parkinson}, a Hilbert-Huang transformation was performed, followed by feature extraction. The Hilbert transform is an explorable option initially and has not been used within the references found, leading to the hypothesis of feature enhancement using this transformation. The reported accuracy is 92.90\% with a sensitivity of 98.90\%, showing paths to explore. A second study developed by Hathaliya et al. (2022) in \cite{hathaliya2022deep}, based on a temporal embedded model GRU-LSTM, demonstrates the potential of a temporal model for predicting Parkinson's disease and non-Parkinson's disease in the "BioStampRC21" dataset \cite{biostamp}, achieving training and validation accuracies of 80.40\% and 74.10\% respectively. The studies conducted by Papadopoulos et al. (2019) and Channa et al. (2022) reported in \cite{papadopoulos2019multiple} and \cite{channa2022parkinson}, respectively, use somewhat different models from those already mentioned. While the first work uses a model called "Multiple Instance Learning," which incorporates at least one attention layer, the second study uses transfer learning under the "AlexNet" architecture, achieving good results. The first study achieves an accuracy of 89.30\% and an F1 score of 85.10\% in a binary prediction of tremor presence and absence compared to the second study of multiclass tremor grade classification for the MJFF Levodopa dataset \cite{daneault2021accelerometer}, with sensitivities of 0.97, 0.93, 0.88, 0.88, and 0.36 for grades 0, 1, 2, 3, and 4 respectively. It is noted that the dataset has a significant class imbalance, which is reflected in the results obtained. 

After commenting on the models found in the studies, we observe the variability among these models, ranging from classical models directly related to feature extraction from signals to deep neural convolutional (CNN) models, temporal models like LSTM and GRU, and ending with Multiple Instance Learning models based on attention layers and AlexNet using transfer learning. All these models are used for classification and tremor recognition, whether binary or multiclass. Although there are differences in the results, the models generally have shown promising results, performing different processing steps that directly influence the models' outcomes. Considering this, the interdependence of a complete signal flow, preprocessing, and the model used for the final results is evident. The next section will discuss hyperparameter selection within the model, as signal reading and preprocessing have already been covered in previous sections.

\subsection{Hyperparameter Selection}

This section analyzes the hyperparameter selection process for the models executed in each study. This selection aims to optimize the model and achieve the best possible result. We will focus our analysis on the methodology for parameter selection when reported in the studies.

In general, as noted in the collection of studies, there are 21 studies where no parameter selection process is reported compared to 12 studies where a methodology for hyperparameter selection is reported regardless of the model used.

Initially, there is no marked pattern for the characteristic of parameter search depending on the model used. According to Skandha et al. (2020) in \cite{skandha20203}, we find models belonging to both the second and third generations where parameter search was not applied to adjust the model. Thus, there is no preference for the model associated with not performing such a parameter search. This does not mean that internal searches were not executed within the experiments conducted in each study. However, the study does not mention performing any parameter search; only a combination of parameters used in the executed model for the obtained result is referenced.

Therefore, we will comment on the studies where we found a parameter selection process, as these will indicate a process that improves models to achieve a better result. For instance, Ranjan et al. (2021) reported in \cite{ranjan2021convolutional} uses the \textbf{GridSearch} method within CNN models to search for the best result based on precision and F1 score metrics for selection. Igene et al. (2023) reported in {} also uses a GridSearch method for a Support Vector Machine (SVM) model. Although reference parameters in the GridSearch are not mentioned, the search is referenced alongside dimensionality reduction using the "Principal Component Analysis" (PCA) method, considering precision metrics. Two final studies performed a parameter search with GridSearch; the first study by Li et al. (2023) reported in \cite{li2023improved} searches for second-generation models, concluding with a better SVM model considering precision and sensitivity metrics. In comparison, a second study by De Araujo et al. (2020) reported in \cite{de2020hand} conducts a GridSearch process for a KNN model and specific combinations in a particular set of second-generation models such as SVM, Linear Regression, among others, concluding with the choice of a KNN model.

A second method for parameter search focuses on executing specific parameter combinations used as options to find the best combination. The first study that considers this method was conducted by Hathaliya et al. (2022), reported in \cite{hathaliya2022deep}, uses the GRU-LSTM embedded temporal architecture, where parameter combinations oriented to the number of iterations, epochs, optimizer, etc., are considered. Similarly, Papadopoulos et al. (2019) reported in \cite{papadopoulos2019detecting} and Ibrahim et al. (2020) reported in \cite{ibrahim2020parkinson} perform parameter selection in a CNN model from a collection of options oriented to the number of layers, number of neurons for each layer, training epochs, etc., achieving good results in both cases for the respective tasks. Finally, Peres et al. (2021) perform iterations on parameter combinations related to the SVM model to find a better model based on the extracted signal features, thus solving the classification task \cite{peres2021discrimination}.

Another parameter selection option is associated with considering related works for a similar task and/or the same dataset used for the conducted study. For instance, Shawen et al. (2020) reported in \cite{shawen2020role} and Kim et al. (2018) reported in \cite{kim2018wrist} consider related studies for parameter selection in the SVM and CNN models, respectively. This does not imply study replication, as significant differences in some processing or feature extraction processes are considered. Still, having a common process and parameter consideration for the model provided a reasonable basis for achieving a good result.

On the other hand, we encounter a method somewhat different from iterations or fixed combinations within a set of model-related parameters such as learning rate, number of neighbors, and number of neurons, among others. A first study by Legaria-Santiago et al. (2022) reported in \cite{legaria2022computer} optimizes hyperparameters using a heuristic function until finding the best model based on precision, recall, and F1 score metrics for a Fuzzy Logic model. A final study by AlMahadin et al. (2021) reported in \cite{almahadin2021parkinson} performs parameter selection using a Bayesian algorithm. It is worth noting that the Bayesian algorithm uses previous evaluations to predict the next set of hyperparameters close to the optimal, functioning as a heuristic algorithm to reach an optimal parameter combination, consequently reducing the number of evaluations required to achieve the best result based on precision metrics. This reference metric can vary but maintains the parameter selection concept.

Finally, we can comment that in a parameter selection process, several viable options are available for executing the process, considering this process block as an essential part of the search for the best possible results directly linked to the model, not executions that directly interfere with signal processing. Even obtaining the best possible outcome for the methodology followed in the experimental part of the study, this result only gains relevance under a direct comparison with related studies' results and/or external validation with data from a different source than the dataset, which will be analyzed in the next section.

\subsection{External Validation and Comparison}

In this section, as mentioned in the last paragraph of the previous section, the analysis will be associated with an external comparison and validation of the results of each study concerning related works. Comparisons generated regarding different models within the same study will not be considered, as the aim is to evaluate the added value relating to studies that have addressed the same or similar tasks as the reference.

A key point of reference is that no study had an external validation, clarifying that this concept is oriented toward testing the implemented model in a real-world setting or on datasets of different origins than those used in the study. The importance of this validation would pave the way for a practical application in a clinical environment, verifying not only the model's robustness but also the performance of intelligent technological tools within an entire health evaluation system. The absence of validation under the declared concept does not invalidate the conclusions reached within the study regarding the model's robustness and/or added value under other methodologies oriented toward non-external validation. Technological development and external validation details must align with sufficient ethics and regulations in the health area, aiming for a safe technology for the patient and stakeholders. Due to this correlation, no studies are oriented toward this characteristic.

On the other hand, 12 studies conducted comparative analyses concerning the results of other related studies. The importance of comparison lies in measuring the level of scientific improvement and the study's contribution to future references correlated with the metrics discussed in previous sections. The studies conducted by Duque et al. (2020) reported in \cite{duque2020angular}, Sigcha et al. (2021) reported in \cite{sigcha2021automatic}, Legaria-Santiago et al. (2022) reported in \cite{legaria2022computer}, Papadopoulos et al. (2019) reported in \cite{papadopoulos2019detecting}, Peres et al. (2021) reported in \cite{peres2021discrimination}, De Araujo et al. (2020) reported in \cite{de2020hand}, Li et al. (2023) reported in \cite{li2023improved}, San Segundo et al. (2020) reported in \cite{san2020parkinson}, Ibrahim et al. (2020) reported in \cite{ibrahim2020parkinson}, Sun et al. (2023) reported in \cite{sun2023parkinson}, Sun et al. (2021) reported in \cite{sun2021tremorsense}, and Kim et al. (2018) reported in \cite{kim2018wrist} conducted direct comparisons with various models from the literature based on related studies addressing the same task, including binary or multiclass classification. The potentialities of each study as an added value can be directly observed under the metrics. In some cases, improvements are also attributed to processing or preprocessing methods to optimize the model, achieving better or very close results.

Finally, we can conclude that for the development of models oriented towards the classification and/or recognition of tremors in Parkinson's disease, there are barriers associated with model validation that do not allow development in a clinical setting, considering the potential of the models discussed in the studies. This is directly related to the existing regulations for developing technological products in health that do not provide the necessary support for such development. Similarly, comparisons suggest good contributions from studies during the development period. These references and methodologies can be evaluated for specific use in the required field without forgetting that there may still be methodological bad practices that need to be corrected.

\subsection{Interpretation of result}

In this section, we discuss the interpretation conducted in each collected study to evaluate whether it highlights potentialities within the results that can serve as references for future studies aimed at similar tasks.

In general, the interpretation of results focuses on metrics such as balanced accuracy, F1 score, recall, and AUC curve to analyze whether the results show significant contributions compared to other studies. The choice of metrics depends on the task performed, meaning whether the developed model executes binary and/or multiclass classification, considering whether the data used is imbalanced.

The only reference where the results obtained are not declared is found in the study conducted by Thorp et al. (2018) reported in \cite{thorp2018monitoring}. The characteristic of not reporting the results obtained does not allow for effective evaluation of such results. Although the obtained metrics are mentioned, the methodology within the experiment, such as data collection, signal filtering, and model implementation, is still being determined. Therefore, it is impossible to interpret these results and the advantages that can be considered for the adopted strategy.

Regarding the remaining studies, all of them present their results by commenting and summarizing the corresponding metrics. These results were also represented using figures showing precision-recall curves, confusion matrices, and the corresponding percentages within the confusion matrices. Such visualizations are important for evaluating false positives and negatives within classification tasks. The significance of these two concepts is heightened within the healthcare field; false positives and false negatives represent patients who were classified as positive but are not and patients who were classified as negative but are positive, respectively. In both cases, there are direct implications for the patient; in the case of a false positive, the patient receives the treatment they do not need, whereas, in the case of a false negative, the patient does not receive necessary treatment, allowing the progression of the condition. It is crucial to consider and minimize the number of false positives and negatives in classification tasks, a topic addressed in the collected studies through F1 score and recall metrics. However, in some cases, these considerations need to be more detailed.

In conclusion, the interpretations of results in all considered studies, except one, show a reasonably comprehensive result interpretation pattern, an important characteristic for the scientific community as these interpretations demonstrate the contributions and advancements made for the task at hand. While this is a key point, there are preceding processes that result in acquisition that have been observed and should be subject to study. Therefore, it is essential to discuss the limitations and future directions based on the descriptions in all the subsections considered within the survey.

\subsection{Limitations and Future Directions}

This section discusses some of the limitations identified in the studies from the authors' perspectives. Limitations should not be seen as insurmountable problems but rather as challenges that can be addressed through new and innovative methods.

\subsubsection{Limitations}

Firstly, one significant limitation identified pertains to the public datasets used, specifically the imbalance in class distribution, which aligns with the prevalence of Parkinson's disease sub-classes. This imbalance poses a challenge as artificial intelligence models heavily depend on data quantity. While there are methodologies to address limited data, their reliability hinges on the original signal quality, making insufficient data a barrier to achieving robust performance, especially for less prevalent sub-classes. As a workaround, some studies transform the problem into a binary classification of the presence or absence of a class grouping multiple sub-classes.

Another notable limitation concerns the characterization of Parkinson's symptomatically. From a computational perspective, various frequency-based characterizations may conflict with each other. Specific frequency bands used for filtering and enhancing features of tremors may not align with physiological definitions, as noted in Figure \ref{fig:figure9}. Many of these references are drawn from related studies cited in the bibliography of each study, suggesting a transfer of considerations that must be scrutinized over time to ensure confidence in the adopted references.

A third limitation concerns the varied methodologies observed across studies. While each study follows steps from signal processing to model consideration, marked differences exist in their implementation. As discussed in previous sections, variations include the methodology for data partitioning into training, testing, and validation sets, where some studies exhibit randomness without respecting temporal dimensionality or introduce biases by including samples from the same patient in different groups. Such methodological differences can lead to misinterpretations of study outcomes.

External factors also present challenges, such as difficulty in sample capture, labeling, and controlling specific data linked to the pathology. These limitations persist and lie beyond the computational control, emphasizing the need for standardized methodologies.

\subsubsection{Future Directions}

Based on the identified limitations, three primary areas should be prioritized for future studies:

\begin{itemize}
    \item Handling Imbalanced Class Distribution: Future studies should focus on developing or adopting standardized methodologies to address imbalanced class distributions. This includes exploring techniques for better representation of less prevalent sub-classes to improve model performance.
    \item Patient-Centric Data Partitioning: Implementing methodologies prioritizing patient-centric data partitioning can mitigate biases introduced during experiments. Ensuring data across different subsets remains coherent and representative is crucial for reliable model evaluation.
    \item Frequency Band Characterization of Tremors: There is a critical need to elucidate the frequency bands characterizing tremors and establish a comparative framework to align computational findings with physiological definitions. This comparative approach can enhance the interpretability and clinical relevance of computational findings.
\end{itemize}

\section{Conclusion}
\label{sec:conclusion}

The studies reviewed in this survey have demonstrated significant advances in recognizing and classifying Parkinson 's-associated tremors. Each study provides valuable contributions, with some offering particularly promising insights that lay the groundwork for future research. The sequence of data loading, preprocessing, appropriate model selection, and hyperparameter tuning have proven crucial in enhancing the accuracy and robustness of the developed models. Methodologies focused on patient-centered approaches show promise in real-world settings, thus garnering greater value in the results and conclusions.

However, significant challenges persist that need addressing. These include dataset imbalances, variations in signal processing methodologies, and the need for more robust external validation to ensure the clinical applicability of the developed models. Future research should overcome these limitations through standardized methods for handling data imbalances, implementing external validation techniques in real clinical environments, and developing more interpretable and explanatory models.

As these models are further refined and validated more broadly, they could significantly impact clinical practice, enabling more precise and personalized diagnoses and facilitating continuous and non-invasive monitoring of Parkinson's patients.

\section{Acknowledgments}
Part of the results presented in this work was obtained through the project ``Hub of Artificial Intelligence in Health and Wellbeing – Viva Bem'', funded by Samsung Eletrônica da Amazônia Ltda., within the scope of the Information Technology Law 8.248/91.

\bibliographystyle{unsrt}  
\bibliography{references}

\end{document}